\documentclass{article}

\PassOptionsToPackage{numbers,compress, sort}{natbib}

\usepackage[preprint]{neurips_2026}

\usepackage{amsmath}
\usepackage{graphicx} 
\usepackage{subcaption}
\usepackage{makecell}
\usepackage{enumitem}

\usepackage[utf8]{inputenc} 
\usepackage[T1]{fontenc}    
\usepackage{hyperref}       
\usepackage{url}            
\usepackage{booktabs}       
\usepackage{amsfonts}       
\usepackage{nicefrac}       
\usepackage{microtype}      
\usepackage{xcolor}         
\usepackage{csquotes}
\usepackage{comment}
\usepackage[capitalise]{cleveref}
\usepackage{algorithm}
\usepackage{algpseudocode}
\usepackage{multirow}
\usepackage{subcaption}

\newcommand{\mypara}[1]{\textbf{#1}.}

\newtheorem{theorem}{Theorem}

\Crefname{figure}{Fig.}{Figs.}
\Crefname{table}{Tab.}{Tabs.}
\Crefname{section}{Sec.}{Secs.}
\crefname{appendix}{App.}{Apps.}
\crefname{equation}{Eq.}{Eqs.}

\title{APM: Evaluating Style Personalization in LLMs with Arbitrary Preference Mappings}

\author{%
  Philipp Spohn\textsuperscript{1} \quad Leander Girrbach\textsuperscript{1,2} \quad Zeynep Akata\textsuperscript{1,2} \\
  \textsuperscript{1}Technical University of Munich, Helmholtz Munich
  \\
  \textsuperscript{2}Munich Center for Machine Learning (MCML)
}

\begin{document}

\maketitle

\begin{abstract}
Typical LLM responses tend to follow a default style, even though users often have distinct preferences regarding tone, verbosity, and formality that they do not explicitly state in their prompts.
Evaluating whether personalization methods can adapt to these implicit preferences is challenging, since users typically provide prompts rather than reference responses, style preferences are not factually verifiable, and reference-free LLM judges may conflate personalization with general response quality.
To address these challenges,  we introduce the Arbitrary Preference Mapping (APM) benchmark, which decouples user attributes (e.g.\ enthusiastic) from response principles (e.g.\ persuasive) via a hidden, randomized mapping $\mathbf{C}$ that maps user attributes to preferences about response traits. Because $\mathbf{C}$ carries no semantic content and is resampled across runs, models cannot exploit stereotypical associations and must infer preferences from conversation history. Using this unbiased evaluation methodology, we adapt retrieval-augmented, prompt-optimization, and routing personalization methods and evaluate them on \texttt{Llama-3.1-8B} and \texttt{Qwen-3.5-27B}. Our results show that routing is the most reliable approach, while RAG only improves with the stronger base LLM, and soft prompt optimization fails to improve significantly over a non-personalized baseline. Our extensive evaluation reveals that in this realistic setting, personalization remains challenging, but our adapted methods show promise.
\end{abstract}

\section{Introduction}
\label{sec:intro}

In a conversation, large language models (LLMs) typically provide responses based on the average user, although in practice, users have different preferences regarding style, such as tone, length, or formality. However, inferring user preferences is a difficult task because we cannot directly ask for them, and prompts often provide limited information. Furthermore, evaluating personalization methods is challenging because it relies on synthetic data and LLM judges, which carry biases. This work addresses both challenges. We investigate how to design personalization methods that rely only on conversation history, i.e.\ without explicit preferences. We also study how to fairly evaluate whether a method achieves true personalization and does not simply exploit semantic priors regarding the relationship between user traits and preferences that may not be true in practice.

Evaluating personalization is difficult because conversational settings offer only indirect supervision. Users consume responses and do not author them, so there are generally no ground-truth texts to compare against.
Stylistic preferences are also typically revealed through interaction rather than stated explicitly, so a personalization system cannot rely on a complete, up-front preference specification.
Existing benchmarks address these limitations in different ways, but each also has weaknesses.
Reference-based benchmarks such as LaMP \citep{salemi2024lamp} and LongLaMP \citep{kumar2024longlamp} use a corpus of user-authored texts and measure how closely a model reproduces the writing of a user. In reality, this signal is not available when users only consume responses. A second line of work targets explicit, verifiable preferences, such as stated dislikes, dietary restrictions, or factual user attributes \citep{zhao2025prefeval, jiang2025personamemv2, jiang2025knowme}, where correctness can be objectively checked, which is not the case for style personalization. Real-world preference datasets such as UltraFeedback \citep{cui2024ultrafeedback}, Anthropic HH \citep{bai2022training}, and HelpSteer3 \citep{wang2025helpsteer3} avoid synthetic preference data, but their annotations are not tied to individual users, so they cannot be used to evaluate style adaptation to specific users.

Reference-free evaluation of non-verifiable preferences therefore relies on LLM-as-a-judge protocols \citep{zheng2023judging}, in which an LLM judge rates whether a response matches a preference profile. However, without objective anchor, we empirically find that LLM judges conflate general response quality with personalization and reward generic well-written outputs regardless of whether they match the target user. We illustrate this by prompting \texttt{Qwen-3.5-27B} \citep{qwen3.5} to \enquote{act as if it knows the user well} without providing any actual user information. In this scenario, a personalization judge rates its response 1.78 vs.\ 0.81 (on a 1--10 scale) compared to a generic baseline. An extended version of this experiment can be found in \cref{app:category_based_scoring} (\cref{tab:judge-bias}).

To address these problems, our benchmark combines three design choices. First, we decompose holistic personalization ratings into narrow, per-principle judgments along stylistic axes (e.g., how elaborate a response is). Each axis describes two ends of a stylistic spectrum rather than a better-versus-worse dimension, which removes the judge's directional bias toward overall response quality, see \cref{app:category_based_scoring} for a more detailed empirical argument in favor of narrow judges. Second, we sample the mapping between user attributes and response principles so that its entries have zero mean. Under this construction, we prove that whatever residual bias the judge still carries cancels out in expectation. Any unpersonalized model achieves an expected reward of zero and a win-rate of exactly 50\%. Third, we keep the mapping hidden and randomize it across evaluation runs, so that semantic priors about which user types typically prefer which styles carry no signal. Any above-baseline performance can then only come from inferring the mapping from the user's conversation history. With our new evaluation methodology, we test RAG-, prompt-optimization-, and routing-based personalization methods, finding that routing-based personalization most reliably exceeds the baseline, with a win-loss ratio of up to 1.79 against the non-personalized model.

Our main contributions are: (1) the Arbitrary Preference Mapping (APM) benchmark, the first reference-free evaluation of implicit style personalization; (2) the adaptation of retrieval-, prompt-optimization-, and routing-based personalization methods to the setting of implicit style personalization from conversation history alone; and (3) an empirical study of routing-based personalization against retrieval- and prompt-optimization-based baselines.

\section{Related Work on Personalization and its Evaluation}
\label{sec:related}

Approaches to LLM personalization can be organized along the following axes: \emph{Retrieval-augmented} methods inject relevant user history into the prompt as few-shot examples or summarized profiles \citep{salemi2024ropg, mysore2024pearl, richardson2023summaryrag}. \emph{Per-user parameter-efficient} methods train a dedicated adapter (typically LoRA) for each individual user \citep{tan2024oppu, tan2024perpcs}. \emph{User-embedding} methods compress interaction history into continuous vectors that condition a frozen base model via soft prompts or cross-attention~\citep{ning2025userllm, liu2025personaplug, hebert2024persoma}. \emph{Decoding-time} methods steer generation via activation engineering or logit manipulation without parameter updates \citep{kim2025drift, zhang2025stylevector}. A closely related family of \emph{group-level} and mixture-of-experts approaches routes queries among pre-trained specialized experts based on user clusters or task types \citep{zhang2025proper, tan2024perpcs, dan2024ptailor, muqeeth2024phatgoose}. See \citep{zhang2024personalizationsurvey, liu2025personalizationsurvey} for extensive surveys of these methods. We adapt retrieval-augmented, user-embedding, and group-level routing methods to our setting, as these families support zero-shot generalization to new users without per-user training.

Existing personalization benchmarks fall into several paradigms. \emph{Reference-based} benchmarks evaluate how well a model can create outputs that look like they come from a given user \citep{salemi2024lamp, kumar2024longlamp, li2025personaconvbench, arocaouellette2025plume}. This type of benchmark requires ground-truth user writings, which do not exist in conversational assistant settings where users consume responses rather than produce target outputs. A second family targets \emph{factual} preferences that can be stated declaratively and verified against a fixed ground-truth attribute, such as dietary restrictions, allergies, or named dislikes \citep{zhao2025prefeval, jiang2025personamemv2, jiang2025knowme, afzoon2024persobench}. Style preferences admit no such check and fall outside the scope of these benchmarks. 

Aspect-based frameworks such as ExPerT~\citep{salemi2025expert}, PREF~\citep{fu2025pref}, and PrefDisco~\citep{li2025prefdisco} decompose holistic personalization judgments into atomic criteria, and PrefDisco in particular applies this to stylistic traits. Our framework takes this idea further and extends it to implicit style preferences inferred from conversation history, combining narrow per-principle classification with a randomized attribute-to-principle mapping that yields a calibrated baseline in which an unpersonalized model's expected reward is exactly zero. Finally, PERG~\citep{okite2025perg} measures the trade-off between following user preferences and maintaining factual correctness, which we include by evaluating all personalization methods on general capability benchmarks in addition to personalization.

\section{Personalization Evaluation through Arbitrary Preference Mappings}

\begin{figure}[t]
    \centering
    \includegraphics[width=\linewidth]{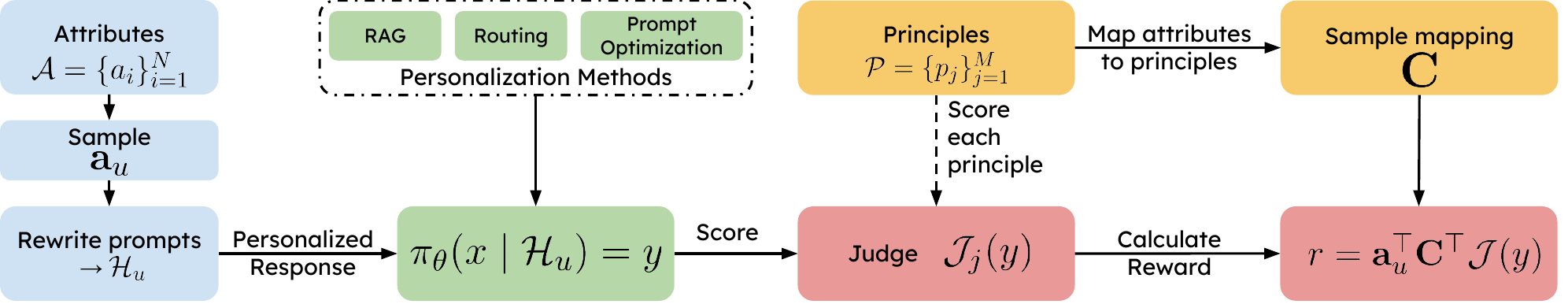}
    \caption{Overview of our APM evaluation method: Sample a mapping matrix $\mathbf{C}$ (upper right) to randomly associate user attributes and response principles, and per-user attribute vectors $\mathbf{a}_u$ (left). Attributes are used to write prompts in the user's style, producing per-user conversation histories $\mathcal{H}_u$ for use in personalization systems $\pi_{\theta}$. A response $y$ is scored by a judge $\mathcal{J}$ along each principle $p_j$. The reward weights the scores according to user preferences defined by $\mathbf{C}$.
    }
    \label{fig:overview}
\end{figure}

We consider a setting where an LLM serves multiple users. Each user $u$ interacts with the assistant across several sessions, which produces a conversation history $\mathcal{H}_u = \{c_1^u, \ldots, c_T^u\}$. The user also has latent style preferences (e.g., preferred tone, length, or formality) that they never state explicitly. A personalized LLM generates a response $y = \pi_\theta(x \mid \mathcal{H}_u)$ to a new query $x$ that satisfies these preferences. Our goal is to evaluate personalization methods that transform standard LLMs into these personalized systems $\pi_\theta$.

The main difficulty is that standard persona-based and LLM-judge evaluations mix two distinct abilities: (i) inferring user preferences from interaction evidence, and (ii) producing responses that a judge considers high-quality regardless of personalization. A model that defaults to long, polite, and cautious outputs can score well on benchmarks without real user-level inference, and one that leans on stereotypical user-to-style associations can appear personalized without observing the user at all. Our evaluation addresses both shortcuts by treating the mapping from user attributes to preferences as an unknown, random function. We design this function so that any unpersonalized model achieves zero expected reward, which cancels judge bias in expectation. Because the mapping additionally carries no semantic meaning, a model also cannot succeed by exploiting semantic priors, but must infer preferences from the conversation history. Below, we show that under mild assumptions about how this mapping is sampled, this setup yields a baseline reward of exactly zero for any unpersonalized policy and any judge. Any positive reward is therefore a clear measurement of successful preference inference. See \cref{fig:overview} for an overview of our evaluation method.

\subsection{Formalizing User Attributes, Response Principles, and the Mapping between them}
\label{sec:formalization}

We model users along two orthogonal axes. First, a set of $N$ user attributes $\mathcal{A} = \{a_i\}_{i=1}^N$ represents stylistic traits expressed in a user's prompts (e.g., $a_i = $ \enquote{enthusiastic}). Each user $u$ is represented by an attribute vector $\mathbf{a}_u \in \{-1, 0, +1\}^N$ with $k$ active (non-zero) entries: $+1$ indicates the user expresses $a_i$, $-1$ indicates the user expresses the opposite (e.g., a dry rather than enthusiastic prompt style), and $0$ indicates the axis is inactive for $u$. Second, a set of $M$ response principles $\mathcal{P} = \{p_j\}_{j=1}^M$ similarly describes stylistic properties of model outputs (e.g., $p_j = $ \enquote{persuasive}). See \cref{app:attribute_selection} and \cref{tab:attributes} for the exact attributes and principles and how they are selected.

\begin{figure}[t]
    \centering
    \includegraphics[width=\linewidth]{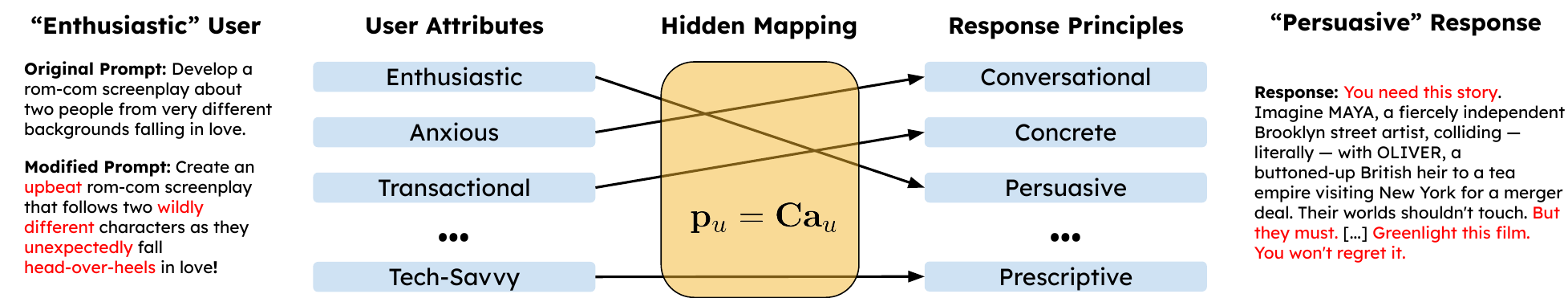}
    \caption{Example of the attribute-to-principle mapping. The attribute vector $\mathbf{a}_u$ is multiplied by $\mathbf{C}$ to yield a latent preference vector $\mathbf{p}_u = \mathbf{C}\mathbf{a}_u$ over response principles. In our experiments, a neutral prompt is adapted to an \textit{enthusiastic} user (left) through upbeat phrasing (highlighted in red). Since $\mathbf{C}$ maps \emph{enthusiastic} to the \emph{persuasive} principle, the target response style is persuasive (right).}
    \label{fig:attribute_mapping}
\end{figure}

A user's latent preferences are defined via a mapping matrix $\mathbf{C} \in \mathbb{R}^{M \times N}$ from attributes to principles:
\begin{equation}
    \mathbf{p}_u = \mathbf{C}\,\mathbf{a}_u \in \mathbb{R}^M.
    \label{eq:preference}
\end{equation}
A positive entry $\mathbf{p}_{u,j} > 0$ indicates that user $u$ prefers responses that follow principle $p_j$. A negative entry indicates a preference for responses that do not follow $p_j$, and the magnitude represents preference strength. Avoiding $p_j$ does not need to correspond to another named principle; it can refer to both the absence and the opposite of $p_j$. To evaluate personalization systems, we sample $\mathbf{C}$ randomly. Any association between user attributes and preferences is therefore learned from training feedback rather than assumed from prior knowledge. See \cref{fig:attribute_mapping} for an example.

\subsection{Deriving the Conditions under which LLM Judge Evaluation is Unbiased}
\label{sec:calibration}

We evaluate personalized responses using an LLM Judge $\mathcal{J}$, where $\mathcal{J}_j(y)$ scores how strongly response $y$ encodes principle $j$. The natural reward under a given mapping $\mathbf{C}$ is
\begin{equation}
    r(u, x, \pi) \;=\; \sum_{j=1}^M \mathbf{p}_{u,j} \cdot \mathcal{J}_j\big(\pi(x \mid \mathcal{H}_u)\big) \;=\; \mathbf{a}_u^\top \mathbf{C}^\top \mathcal{J}\big(\pi(x \mid \mathcal{H}_u)\big).
    \label{eq:reward}
\end{equation}
We observe that $\mathcal{J}$ can be systematically biased. For instance, it could reward longer or more formal responses regardless of content, which can inflate reward for any system that produces outputs the judge generally favors, even without personalization. The following result shows that, under suitable sampling of $\mathbf{C}$, these biases have no effect on the baseline expected reward.

\begin{theorem}
\label{lemma:baseline}
Let $\pi_0$ be a policy that does not condition on the user's history $\mathcal{H}_u$, and let $y_0 = \pi_0(x)$. Assume: (i) $\mathbb{E}[\mathbf{C}_{ji}] = 0$ for all $(j, i)$ and (ii) $\mathbf{C}$ is independent of $(\mathbf{a}_u, x, y_0)$. Then $\mathbb{E}[r(u, x, \pi_0)] = 0$ for any judge $\mathcal{J}$.
\end{theorem}

\mypara{Proof}
By \eqref{eq:reward} and linearity of expectation,
\begin{equation}
    \mathbb{E}[r(u, x, \pi_0)] = \sum_{j,i} \mathbb{E}\big[\mathbf{C}_{ji} \cdot \mathbf{a}_{u,i} \cdot \mathcal{J}_j(y_0)\big].
\end{equation}
Intuitively, sign-randomizing the entries of $\mathbf{C}$ ensures that, across sampled mappings, each response principle is equally likely to be preferred or dispreferred by the user. Any systematic judge bias toward or against a given principle is then multiplied by a zero-mean coefficient and cancels in expectation.
Formally, by assumption (ii), $\mathbf{C}_{ji}$ is independent of the remaining factors, so each summand factors as $\mathbb{E}[\mathbf{C}_{ji}] \cdot \mathbb{E}[\mathbf{a}_{u,i}\,\mathcal{J}_j(y_0)]$, and by (i) each term vanishes. The judge's bias $\mathbb{E}[\mathcal{J}_j(y_0)]$ can be arbitrary: it is absorbed by the zero-mean factor and does not contribute. In \cref{sec:supp:baselinelemma}, we additionally show that the baseline win-rate is exactly $50\%$ if $\mathbf{C}\mathbf{a}_u$ is equal in distribution to $-\mathbf{C}\mathbf{a}_u$.

Assumption (i) effectively requires independent sign randomization when sampling $\mathbf{C}$, i.e.\ we can choose the absolute values of entries arbitrarily, but their signs must be distributed uniformly. Assumption (ii) requires only that $\mathbf{C}$ is sampled independently of user attributes and prompts.

\section{Personalization Methods: RAG, Routing, and Prompt Optimization}
\label{sec:systems}

\begin{figure}[t]
\centering
\includegraphics[width=\linewidth]{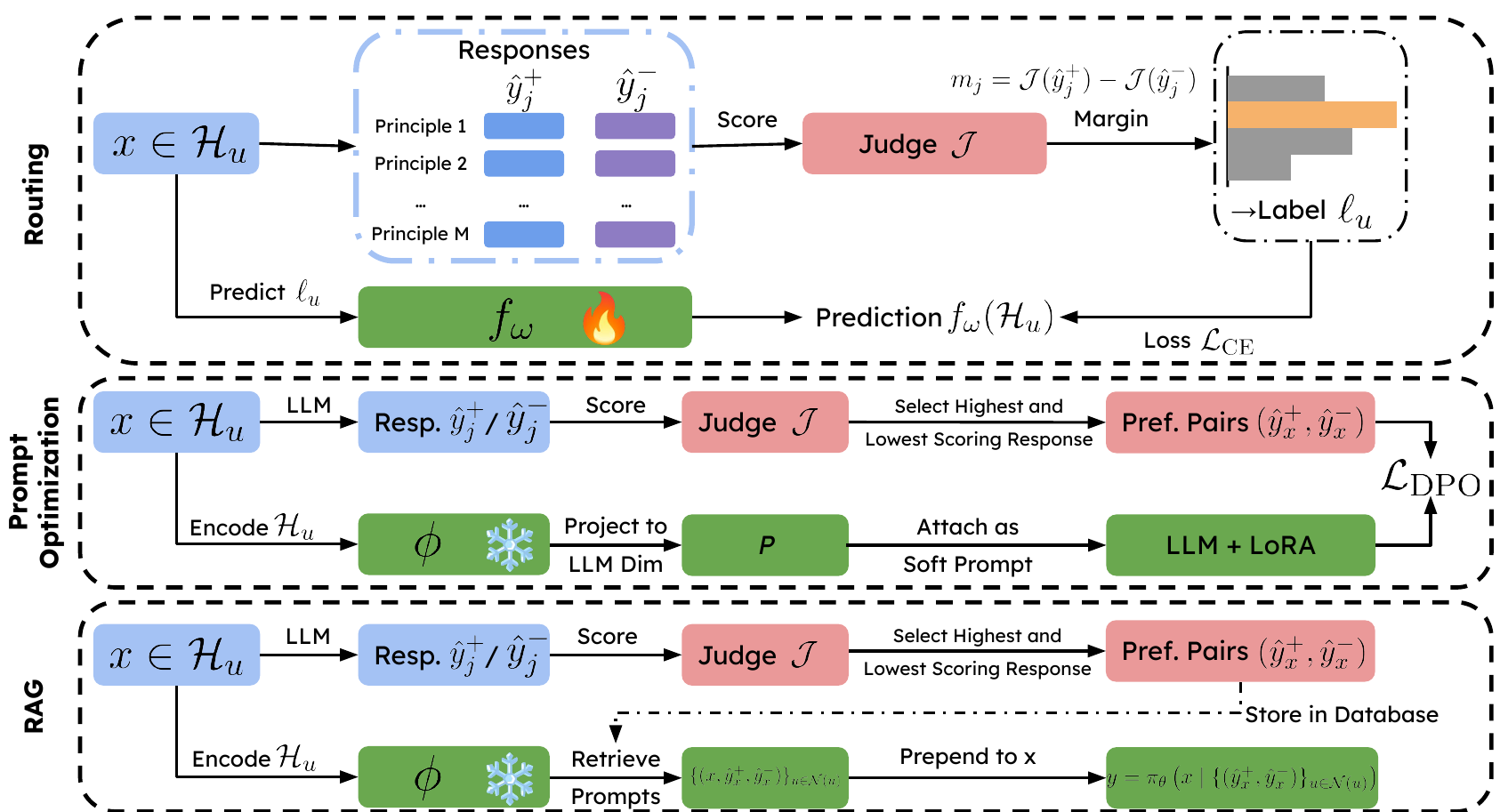}
\caption{Personalization methods. \emph{Routing} (top) trains a classifier 
$f_\omega$ to predict a (principle, direction) label $\ell_u$ from $\mathcal{H}_u$, 
which is converted into a style instruction. \emph{Prompt Optimization} (middle) 
projects $\phi(\mathcal{H}_u)$ into a soft prompt $\mathbf{e}_u = P\,\phi(\mathcal{H}_u)$ 
and trains $P$ jointly with a LoRA adapter via DPO on preference pairs 
$(\hat{y}_x^+, \hat{y}_x^-)$. \emph{RAG} (bottom) retrieves the $k$ nearest 
training users by $\phi$-similarity and conditions generation on either their 
exemplar pairs or LLM-generated preference summaries.}
\label{fig:methods}
\vspace{-10pt}
\end{figure}

We evaluate three personalization methods (see \cref{fig:methods}): Retrieval-Augmented Generation (RAG), Prompt Optimization, and Routing, alongside a \textit{non-personalized baseline} that generates $y = \pi_\theta(x)$ from the prompt alone. All three methods share a common interface: given the conversation history $\mathcal{H}_u$ of user $u$ and a new prompt $x$, methods produce a response $y$, and at training time they can query the LLM-simulated user for a scalar score on any candidate response. The methods do not know the user attribute vectors $\mathbf{a}_u$ or the mapping $\mathbf{C}$, and the set of user attributes itself is hidden. However, the response principles are known. This reflects a realistic deployment setting, i.e.\ we cannot assume any particular user model, but the stylistic axes used to steer the LLM must be specified in advance.

\mypara{RAG-Based Personalization}
The retrieve-then-generate framework \citep{lewis2020retrieval} is widely used for personalization \citep{salemi2024ropg, richardson2023summaryrag}. Here, we retrieve similar users from a labeled pool, infer the style of a new user from the retrieved similar users, and condition generation on the inferred style.

For each training user $u$ and prompt $x \in \mathcal{H}_u$, we generate one candidate response per principle and one for its negation, i.e. $2M$ responses in total. Then, we score each response with the simulated user, retaining the best- and worst-scoring responses $\hat{y}_x^+$ and $\hat{y}_x^-$ as a contrastive pair. Given a test user $u^*$, we retrieve the $k=3$ nearest training users by text encoder ($\phi = $ \texttt{BGE-M3} \citep{chen2024bge}) similarity:
\begin{equation}
    \mathcal{N}(u^*) \;=\; \text{top-}k_{u \in \mathcal{U}_{\text{train}}}\,
    \cos\!\big(\phi(\mathcal{H}_{u^*}),\; \phi(\mathcal{H}_u)\big).
\end{equation}
The \emph{exemplar-based} variant injects $\{(x, \hat{y}_x^+, \hat{y}_x^-)\}_{u \in \mathcal{N}(u^*)}$ directly as in-context examples. However, since $\phi$ often focuses on topic overlap in prompts rather than stylistic similarity, retrieved users may not share the test user's preferences. To address this, we create a \emph{summary-based} variant. Here, we precompute two LLM-generated summaries per training user: a style summary $\sigma_u^{\text{style}}$ of $\mathcal{H}_u$ and a preference summary $\sigma_u^{\text{pref}}$ describing the delta between $\hat{y}_x^+$ and $\hat{y}_x^-$. We then retrieve by computing similarities of style descriptions, i.e.\ using $\phi(\sigma_u^{\text{style}})$, but inject $\{\sigma_u^{\text{pref}}\}_{u \in \mathcal{N}(u^*)}$, i.e.\ the preference summaries, into the generation prompt to guide responses.

\mypara{Prompt Optimization-Based Personalization}
User-embedding methods \citep{ning2025userllm, liu2025personaplug, hebert2024persoma} compress a user's conversation history, i.e.\ the user-authored prompts, into a continuous vector that conditions the LLM. We adapt this idea by mapping the user's conversation embedding to a soft prompt that steers generation toward the user's inferred preferences.

We reuse the preference pairs $(\hat{y}_x^+, \hat{y}_x^-)$ from the RAG training step. Given the frozen text encoder $\phi = $ \texttt{BGE-M3} \citep{chen2024bge}, we learn a linear projector $P$ that maps the user's conversation history embedding to a soft-prompt token embedding:
\begin{equation}
    \mathbf{e}_u = P\,\phi(\mathcal{H}_u) \in \mathbb{R}^{d_t},
\end{equation}
where $d_t$ is the embedding dimension of the LLM. Following \citep{lester2021power,xiang2021prefix}, we prepend $\mathbf{e}_u$ to the token embeddings of the input prompt $x$ and jointly train $P$ and a LoRA adapter on the base LLM via DPO \citep{rafailov2024direct} on the pairs $(\hat{y}_x^+, \hat{y}_x^-)$. At inference, we encode the test user's history with $\phi$, project it through $P$, and prepend the resulting soft prompt to the query. Overall, the projector and LoRA adapter learn jointly to translate stylistic signals in $\phi(\mathcal{H}_u)$ into generation behavior according to user preferences.

\mypara{Routing-Based Personalization}
RAG and prompt optimization condition the LLM on history-derived context, such as retrieved exemplars or a soft prompt. In contrast, routing separates preference inference from generation. A classifier predicts a discrete style label from $\mathcal{H}_u$, and the LLM is conditioned on the predicted style by injecting it into the system prompt through a suitable instruction.

For each training user $u$ and principle $j$, we generate two candidate responses using opposite style instructions: $\hat{y}_j^+$ follows an instruction to adhere to principle $j$ (e.g.\ \enquote{be persuasive}) and $\hat{y}_j^-$ follows an instruction to violate it (e.g.\ \enquote{avoid being persuasive}). We compute the within-principle margin $m_j = \mathcal{J}_j(\hat{y}_j^+) - \mathcal{J}_j(\hat{y}_j^-)$ and assign $u$ the label
\begin{equation}
\label{eq:marginbasedlabeling}
\ell_u \;=\; \big(j^*,\, \text{sign}(m_{j^*})\big),
\qquad j^* = \arg\max_j |m_j|,
\end{equation}
which represents the principle with the strongest preference signal and a sign indicating whether the user prefers $\hat{y}_{j^*}^+$ ($+1$) or $\hat{y}_{j^*}^-$ ($-1$).

Using margins within responses for a single principle makes the decision robust to absolute score differences across principles. A ModernBERT \citep{modernbert} classifier $f_\omega$ is trained to predict $\ell_u$ from $\mathcal{H}_u$. At inference, the predicted response principle $f_\omega(\mathcal{H}_{u^*})$ is converted into a style instruction for the system prompt. To establish an upper bound on this method, we also evaluate a \emph{routing oracle} that bypasses the classifier and directly conditions the LLM on the user's true top principle $j^* = \arg\max_j |\mathbf{p}_{u,j}|$ and the sign $\text{sign}(\mathbf{p}_{u,j^*})$. The oracle separates the impact of style-instruction conditioning from the difficulty of inferring preferences from $\mathcal{H}_u$.

\section{Experiments}

\mypara{Metrics}
We report two metrics on APM: the \textit{$\Delta$ Score}, i.e.\ the per-user difference between the personalized and non-personalized response scores, and the \textit{W/L Ratio}, i.e.\ the ratio of users where the personalized response strictly outscores the non-personalized one to users where it strictly scores lower than the non-personalized one (ties excluded). Additionally, we measure side effects on general capabilities using
MMLU-Pro \citep{wang2024mmlu}, TruthfulQA \citep{lin2022truthfulqa}, and
demographic fairness on BBQ (age) \citep{parrish2022bbq}.

\mypara{Experimental Setup}
We evaluate \texttt{Llama-3.1-8B} \citep{grattafiori2024llama} and \texttt{Qwen-3.5-27B} \citep{qwen3.5} as base LLMs across all personalization methods, with \texttt{NVIDIA-Nemotron-3-Super} \citep{nvidia_nemotron_3_2025} generating candidate responses and \texttt{gpt-oss-120b} \citep{openai2025gptoss120bgptoss20bmodel} both as judge (see \cref{sec:supp:judgeselection} for a comparison) and rewriting UltraChat \citep{ding2023ultrachat} prompts into per-user histories (see \cref{sec:supp:promptrewriting}). Unless stated otherwise, we use $N=M=10$, $k=1$ active attribute per user, $4000$ training and $1000$ test users, $2$ conversation turns per user, and average results over $10$ random signed permutation mappings $\mathbf{C}$. See \cref{sec:supp:experimentaldetails} for full details.

\subsection{Evaluating Personalization Methods on our APM Benchmark}
\label{sec:experiments:main}

Results are in \cref{tab:main}. Routing is consistently the strongest method across both base LLMs. On \texttt{Qwen-3.5-27B} with one active attribute, it reaches a W/L of $1.79$ and $\Delta = +1.11$. This is more than four times the gain of the next-best method, RAG-Summary ($\Delta = +0.26$). The advantage holds with two active attributes ($1.68$ vs.\ $1.12$) and on the smaller \texttt{Llama-3.1-8B} model ($1.40$ at $k=1$, $1.27$ at $k=2$). On the smaller model, the absolute margins shrink, but the ranking does not change. 

At the same time, absolute W/L values range between $1.0$ and $1.8$ for all non-oracle methods. This indicates that our methods achieve substantial but not perfect personalization. This reflects the realistic difficulty of implicit style personalization from short conversation histories, as we study in this work. For example, many prompts, such as short factual questions, allow for very limited stylistic variation in answers, and cues about user style can be sparse. Finally, the non-personalized baseline occasionally produces responses that match a user's arbitrary preferences by chance. The gap to the routing oracle ($1.79$ vs.\ $2.88$ on Qwen, $1.40$ vs.\ $1.64$ on Llama) further indicates that the bottleneck is the router's ability to infer preferences from $\mathcal{H}_u$ rather than the routing method itself. We investigate this further in \cref{sec:experiments:routingablation}.

The two RAG variants improve over the non-personalized baseline on \texttt{Qwen-3.5-27B} ($1.17$ and $1.16$) but are flat on \texttt{Llama-3.1-8B} ($0.99$ and $0.98$ at $k=1$). This suggests that retrieval-based personalization is limited by the base LLM's ability to interpret the user and preference information in retrieved examples, rather than by the retriever itself. This is consistent with the flat scaling with respect to the number of training users in \cref{tab:scaling_train}.

Prompt optimization fails to improve over the baseline ($1.00$ W/L, $\Delta = -0.01$ on Llama), which we attribute to noisy signal from DPO and the method must learn to project frozen \texttt{BGE-M3} embeddings into a soft prompt that steers generation. Because these embeddings are dominated by semantic rather than stylistic structure, this learning problem is too weakly supervised to converge on a useful style representation in our setting.

Our evaluation of general capabilities (MMLU-Pro) and demographic fairness (BBQ) shows that all personalization methods perform slightly worse than the non-personalized baseline. The most extreme result is on BBQ ($-0.141$ for Routing on Qwen, $-0.082$ for the routing oracle). This effect is larger on Qwen than on Llama, which suggests that style conditioning interacts more strongly with the capabilities of the larger model. However, these differences are generally small, and reducing them further remains a goal for future personalization methods.

\begin{table}[t]
\centering
\resizebox{\linewidth}{!}{%

\begin{tabular}{@{}ll cc cc ccc@{}}
\toprule
& & \multicolumn{2}{c}{\textbf{APM (1 attr)}} & \multicolumn{2}{c}{\textbf{APM (2 attr)}} & \multicolumn{3}{c}{\textbf{Robustness ($\Delta$, 1 attr)}} \\
\cmidrule(lr){3-4} \cmidrule(lr){5-6} \cmidrule(lr){7-9}
\textbf{Base LLM} & \textbf{Method} & W/L $\uparrow$ & $\Delta$ Score $\uparrow$ & W/L $\uparrow$ & $\Delta$ Score $\uparrow$ & MMLU-Pro & TruthfulQA & BBQ \\
\midrule
\multirow{6}{*}{Llama-3.1-8B}
& Non-personalized   & $1.01_{\pm 0.05}$ & $+0.01_{\pm 0.06}$ & $1.02_{\pm 0.06}$ & $+0.01_{\pm 0.04}$ & \phantom{-} 0.335 & \phantom{-} 0.504 & \phantom{-} 0.929 \\
& RAG-Exemplar       & $0.99_{\pm 0.07}$ & $+0.01_{\pm 0.09}$ & $1.04_{\pm 0.07}$ & $+0.03_{\pm 0.09}$ & $-$0.020 & +0.054 & $-$0.049 \\
& RAG-Summary        & $0.98_{\pm 0.05}$ & $-0.02_{\pm 0.07}$ & $1.02_{\pm 0.05}$ & $+0.02_{\pm 0.06}$ & $-$0.016 & +0.007 & $-$0.025 \\
& Prompt Optim.     & $1.00_{\pm 0.06}$ & $-0.01_{\pm 0.15}$ & $1.07_{\pm 0.08}$ & $+0.09_{\pm 0.15}$ & $-$0.033 & +0.127 & $-$0.050 \\
& Routing            & $\mathbf{1.40}_{\pm 0.09}$ & $\mathbf{+0.65}_{\pm 0.21}$ & $\mathbf{1.27}_{\pm 0.14}$ & $\mathbf{+0.32}_{\pm 0.14}$ & $-$0.033 & $-$0.047 & $-$0.035 \\
& Routing (oracle)   & $1.64_{\pm 0.15}$ & $+1.00_{\pm 0.13}$ & $1.72_{\pm 0.10}$ & $+0.73_{\pm 0.08}$ & $-$0.026 & $-$0.042 & $-$0.064 \\
\midrule
\multirow{5}{*}{Qwen-3.5-27B}
& Non-personalized   & $1.04_{\pm 0.08}$ & $+0.02_{\pm 0.04}$ & $1.01_{\pm 0.07}$ & $+0.01_{\pm 0.05}$ & \phantom{-} 0.575 & \phantom{-} 0.810 & \phantom{-} 0.929 \\
& RAG-Exemplar       & $1.17_{\pm 0.09}$ & $+0.22_{\pm 0.08}$ & $1.08_{\pm 0.09}$ & $+0.09_{\pm 0.06}$ & $-$0.003 & $-$0.042 & $-$0.046 \\
& RAG-Summary        & $1.16_{\pm 0.08}$ & $+0.26_{\pm 0.16}$ & $1.12_{\pm 0.08}$ & $+0.17_{\pm 0.08}$ & $-$0.023 & $-$0.064 & $-$0.072 \\
& Routing            & $\mathbf{1.79}_{\pm 0.19}$ & $\mathbf{+1.11}_{\pm 0.27}$ & $\mathbf{1.68}_{\pm 0.18}$ & $\mathbf{+0.71}_{\pm 0.16}$ & $-$0.045 & $-$0.105 & $-$0.141 \\
& Routing (oracle)   & $2.88_{\pm 0.27}$ & $+1.93_{\pm 0.18}$ & $2.68_{\pm 0.23}$ & $+1.35_{\pm 0.12}$ & $-$0.025 & $-$0.062 & $-$0.082 \\
\bottomrule
\end{tabular}
}
\caption{Main results across base LLMs. \textbf{APM} columns report W/L ratio and $\Delta$ score on the random-preference-mapping benchmark with 1 or 2 active user attributes (mean $\pm$ standard deviation across sampled mappings). \textbf{Robustness} reports the change in MMLU-Pro, TruthfulQA, and BBQ accuracy relative to each model's non-personalized baseline. APM results are averaged over 10 randomly sampled preference mappings. Bold marks the best non-oracle method per column.}
\label{tab:main}
\vspace{-15pt}
\end{table}

\subsection{Ablating Labeling Strategies of the Routing-Based Personalization Method}
\label{sec:experiments:routingablation}

The gap between routing-based personalization and the routing oracle on \texttt{Qwen-3.5-27B} (W/L $1.79$ vs.\ $2.88$ in \cref{tab:main}) indicates that the labels used for the classifier are the bottleneck, not the routing method itself. We therefore ablate how different label strategies affect performance.

\mypara{Alternative Labeling Strategies}
Recall from \cref{eq:marginbasedlabeling} that our default \emph{margin-based} labeling strategy assigns each user a preference label $\ell_u = (j^*,\, \text{sign}(m_{j^*}))$ with $j^* = \arg\max_j |m_j|$ and within-principle margin $m_j = \mathcal{J}_j(\hat{y}_j^+) - \mathcal{J}_j(\hat{y}_j^-)$. We compare this strategy against three alternatives. \emph{Two-sided argmax} uses the same response pairs $(\hat{y}_j^+, \hat{y}_j^-)$ but treats all $2M$ candidates as independent options, picking the (principle, direction) with the highest raw score, $\ell_u = \arg\max_{(j,d) \in [M] \times \{+,-\}} \mathcal{J}_j(\hat{y}_j^d)$. This removes the effect of the within-principle margin. \emph{One-sided argmax} halves the generation cost by computing only $\hat{y}_j^+$. It then selects the principle whose score deviates most from the rating-scale midpoint $c$ (e.g.\ $5.5$ on our 1--10 scale): $j^* = \arg\max_j |\mathcal{J}_j(\hat{y}_j^+) - c|$ with sign $\operatorname{sign}(\mathcal{J}_j(\hat{y}_{j^*}^+) - c)$. One limitation is that this requires the judge score distribution to be symmetric around $c$. \emph{Regression} drops the classification objective and minimizes the mean squared error against the full $2M$-dimensional judge-score vector $\mathbf{s}_u = [\mathcal{J}_j(\hat{y}_j^+),\,\mathcal{J}_j(\hat{y}_j^-)]_{j=1}^M$. This preserves the magnitude of each principle's preference instead of collapsing it to a single label and direction.

\begin{figure}[t]
\centering
\includegraphics[width=\linewidth]{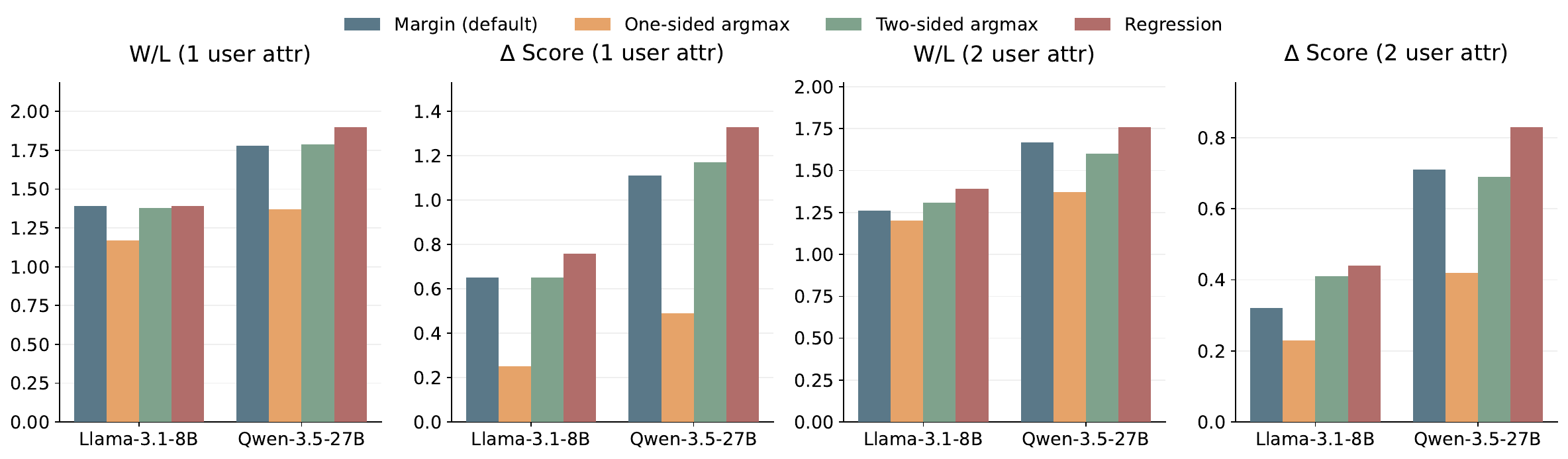}
\caption{Comparison of routing label strategies (margin, one-sided argmax, two-sided argmax, regression) across base LLMs and number of active user attributes. Regression yields the highest W/L and $\Delta$ Score throughout; one-sided argmax is consistently weakest.}
\label{fig:routing_ablation}
\end{figure}

\textbf{Results} are in \cref{fig:routing_ablation}.
Regression is the strongest strategy. It reaches a W/L of $1.90$ and $\Delta = +1.33$ on \texttt{Qwen-3.5-27B} (1 attribute per user) and $1.39$/$+0.76$ on \texttt{Llama-3.1-8B}, and results using two active user attributes confirm this. This means that richer supervision improves performance, but it requires continuous judge scores instead of pairwise comparisons or rankings, which are easier to obtain. Two-sided argmax performs similarly to margin-based labeling (e.g.\ $1.79$/$+1.17$ vs.\ $1.79$/$+1.11$ on \texttt{Qwen-3.5-27B}, 1 attribute per user), which suggests that within-principle normalization is not strictly necessary with a suitable judge model. One-sided argmax yields the weakest performance. This drop is particularly pronounced on the stronger \texttt{Qwen-3.5-27B} model, where $\Delta$ falls by $-0.62$ ($+0.49$ vs.\ $+1.11$), compared to a drop of $-0.40$ on \texttt{Llama-3.1-8B} ($+0.25$ vs.\ $+0.65$) relative to margin-based labeling. This shows that both $\hat{y}_j^+$ and $\hat{y}_j^-$ responses are necessary to learn user preferences effectively. Overall, our results confirm the findings from \cref{tab:main}. Closing the gap to the routing oracle requires richer label supervision, like full score vectors, rather than a different personalization method. Therefore, practitioners face a trade-off between richer supervision with higher performance and simpler labels with reduced performance that are easier to obtain.

\subsection{Ablating Components of the APM Benchmark}

\mypara{Scaling Behavior}
Here, we study how performance scales with the amount of labeled supervision (\textit{number of training users} in $\{500, 1000, 2000, 4000\}$) and the amount of per-user evidence (\textit{conversation-history length} in $\{1, 2, 4\}$ prompts). Results for the number of training users are in \cref{tab:scaling_train}. Routing performance scales monotonically from $\Delta = +0.38$ (W/L $1.23$) at $500$ users to $+1.11$ (W/L $1.79$) at $4000$. This suggests that the classifier benefits significantly from additional users. RAG variants remain flat across settings, e.g.\ RAG-Exemplar moves from $+0.17$ to $+0.22$ (W/L $1.13 \rightarrow 1.17$). This suggests that their bottleneck is not the number of users available for retrieval, but the ability of the LLM to extract style information from retrieved exemplars or summaries. The oracle remains constant at $\Delta = +1.93$ (W/L $2.88$) because it does not depend on the training set.

All learning-based methods improve with more conversations, see \cref{tab:scaling_history} for the results. Routing benefits the most, nearly doubling from $\Delta = +0.76$ (W/L $1.52$) at one conversation to $+1.50$ (W/L $2.21$) at four and closing roughly two-thirds of the gap to the oracle ($+1.93$, W/L $2.88$). RAG-Summary scales clearly as well ($\Delta$: $+0.05 \rightarrow +0.26 \rightarrow +0.40$; W/L: $1.03 \rightarrow 1.16 \rightarrow 1.23$), while RAG-Exemplar does not ($+0.18 \rightarrow +0.22 \rightarrow +0.23$). This indicates that the LLM-generated style summary is the component that absorbs the additional signal, not the raw retrieved exemplars. Together with the results on scaling with the number of users, both more users and richer signals improve performance significantly. However, even at $4000$ training users and four conversations per user, routing approaches but does not match the oracle.

\begin{table}[t]
\centering
\small
\begin{subtable}[t]{0.55\textwidth}
    \centering
    \resizebox{\linewidth}{!}{
    \begin{tabular}{@{}l cc cc cc cc@{}}
    \toprule
    & \multicolumn{2}{c}{\textbf{500}} & \multicolumn{2}{c}{\textbf{1000}} & \multicolumn{2}{c}{\textbf{2000}} & \multicolumn{2}{c}{\textbf{4000}} \\
    \cmidrule(lr){2-3} \cmidrule(lr){4-5} \cmidrule(lr){6-7} \cmidrule(lr){8-9}
    \textbf{Method} & $\Delta$ & W/L & $\Delta$ & W/L & $\Delta$ & W/L & $\Delta$ & W/L \\
    \midrule
    RAG-Exemplar     & +0.17 & 1.13 & +0.13 & 1.10 & +0.15 & 1.10 & +0.22 & 1.17 \\
    RAG-Summary      & +0.16 & 1.10 & +0.18 & 1.16 & +0.24 & 1.16 & +0.26 & 1.16 \\
    Routing          & +0.38 & 1.23 & +0.49 & 1.32 & +0.87 & 1.56 & +1.11 & 1.79 \\
    Routing (oracle) & +1.93 & 2.88 & +1.93 & 2.88 & +1.93 & 2.88 & +1.93 & 2.88 \\
    \bottomrule
    \end{tabular}
    }
    \caption{Scaling with the number of training users.}
    \label{tab:scaling_train}
\end{subtable}%
\hfill
\begin{subtable}[t]{0.44\textwidth}
    \centering
    \resizebox{\linewidth}{!}{
    \begin{tabular}{@{}l cc cc cc@{}}
    \toprule
    & \multicolumn{2}{c}{\textbf{1 conv.}} & \multicolumn{2}{c}{\textbf{2 conv.}} & \multicolumn{2}{c}{\textbf{4 conv.}} \\
    \cmidrule(lr){2-3} \cmidrule(lr){4-5} \cmidrule(lr){6-7}
    \textbf{Method} & $\Delta$ & W/L & $\Delta$ & W/L & $\Delta$ & W/L \\
    \midrule
    RAG-Exemplar     & +0.18 & 1.12 & +0.22 & 1.17 & +0.23 & 1.14 \\
    RAG-Summary      & +0.05 & 1.03 & +0.26 & 1.16 & +0.40 & 1.23 \\
    Routing          & +0.76 & 1.52 & +1.11 & 1.79 & +1.50 & 2.21 \\
    Routing (oracle) & +1.93 & 2.88 & +1.93 & 2.88 & +1.93 & 2.88 \\
    \bottomrule
    \end{tabular}
    }
    \caption{Scaling with history length.}
    \label{tab:scaling_history}
\end{subtable}
\caption{Scaling with training-set size and history length (\texttt{Qwen-3.5-27B}, $N=10$, 1 attr.\ per user).}
\label{tab:both_scaling}
\vspace{-20pt}
\end{table}

\mypara{Benchmark Difficulty}
We evaluate how the size of the attribute and principle pools affects 
personalization performance. We scale both jointly, $N = M \in \{2, 4, 6, 8, 10\}$, which keeps $\mathbf{C}$ square. Increasing this number makes selecting the correct principle more difficult and increases the chance that multiple principles lead to similar responses. Results in \cref{tab:scaling_pool} show that, as expected, personalization becomes harder as $N=M$ grows. Routing remains the strongest method, but the gap to the oracle widens. This reflects increasing uncertainty about the correct response principle when the amount of user interaction remains constant but the number of principles increases.

\begin{table}[t]
\centering
\small
\begin{tabular}{@{}l cc cc cc cc cc@{}}
\toprule
& \multicolumn{2}{c}{\textbf{$N=2$}} & \multicolumn{2}{c}{\textbf{$N=4$}} & \multicolumn{2}{c}{\textbf{$N=6$}} & \multicolumn{2}{c}{\textbf{$N=8$}} & \multicolumn{2}{c}{\textbf{$N=10$}} \\
\cmidrule(lr){2-3} \cmidrule(lr){4-5} \cmidrule(lr){6-7} \cmidrule(lr){8-9} \cmidrule(lr){10-11}
\textbf{Method} & $\Delta$ & W/L & $\Delta$ & W/L & $\Delta$ & W/L & $\Delta$ & W/L & $\Delta$ & W/L \\
\midrule
RAG-Exemplar     & +0.75 & 1.46 & +0.33 & 1.22 & +0.37 & 1.21 & +0.20 & 1.13 & +0.22 & 1.17 \\
RAG-Summary      & +0.38 & 1.20 & +0.46 & 1.25 & +0.58 & 1.37 & +0.44 & 1.24 & +0.26 & 1.16 \\
Routing          & +2.47 & 4.61 & +2.32 & 3.78 & +1.91 & 2.58 & +1.39 & 2.18 & +1.11 & 1.79 \\
Routing (oracle) & +2.43 & 5.22 & +2.29 & 3.23 & +2.12 & 2.85 & +2.04 & 2.89 & +1.93 & 2.88 \\
\bottomrule
\end{tabular}
\caption{Scaling with the joint pool size $N=M$ (\texttt{Qwen-3.5-27B}, 1 attribute per user). Larger $N=M$ corresponds to a harder $2M$-way classification problem.}
\label{tab:scaling_pool}
\vspace{-15pt}
\end{table}

\begin{figure}[t]
\centering
\includegraphics[width=\linewidth]{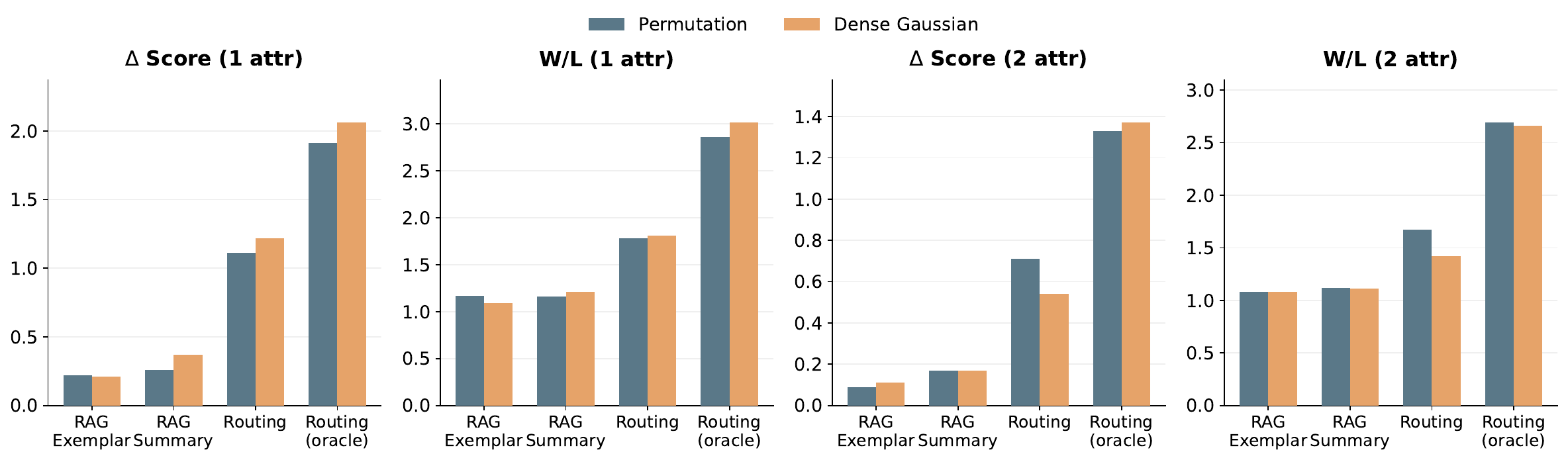}
\caption{$\Delta$ Score and W/L on \texttt{Qwen-3.5-27B} when $\mathbf{C}$ is sampled as a signed permutation (default) vs.\ a dense Gaussian matrix, for 1 and 2 active attributes. Method rankings and magnitudes are largely preserved, indicating that findings do not depend on the structural choice of $\mathbf{C}$.}
\label{fig:mapping_dist}
\vspace{-10pt}
\end{figure}

\mypara{Sampling of Preference Mapping $\mathbf{C}$}
Finally, in \cref{fig:mapping_dist} we show that our results do not depend strongly on the specific structure of the mapping. Concretely, we compare replacing the signed permutation matrix with a dense Gaussian matrix (i.e., iid entries drawn from $\mathcal{N}(0, 1)$), which couples each attribute to all principles with varying strength. Both satisfy the zero-mean condition of \cref{lemma:baseline}, so the baseline win-rate of $50\%$ holds in either case. The qualitative size of effects is largely unchanged, which confirms that our findings are not an artifact of the specific structure of $\mathbf{C}$.

\section{Conclusion}
\label{sec:conclusion}

In this paper, we introduce the Arbitrary Preference Mapping (APM) benchmark to evaluate implicit style personalization in LLMs. Unlike previous work, APM targets a realistic and practical setting where information is limited. In this challenging case, the system must infer a user style from a small number of prompts without explicit preference statements or user-authored reference texts. This setting is more difficult than benchmarks that rely on user-authored data, stated preferences, or semantic links between persona and style. Therefore, it requires an evaluation protocol that does not mix personalization with general response quality.

To measure true personalization instead of exploitation of semantic priors or LLM-judge biases, APM uses a hidden, randomized mapping $\mathbf{C}$ between user attributes and response principles that yields a calibrated baseline (\cref{lemma:baseline}). Our evaluation framework assumes a fixed set of response principles, but the personalization methods only observe $\mathcal{H}_u$ and do not make assumptions about the underlying user model, rendering them flexible and relevant in practical deployment.

In this setting, we also adapt three types of personalization methods to work without explicit preferences or reference outputs: retrieval-augmented generation, prompt optimization based on user embeddings, and routing. Our experiments show that routing-based personalization most reliably exceeds the baseline, reaching a W/L ratio of $1.79$ on \texttt{Qwen-3.5-27B}. RAG variants improve only with a stronger model, and prompt optimization with a frozen encoder fails to outperform the non-personalized baseline. The gap between our best method and the routing oracle ($1.79$ vs.\ $2.88$) shows that the main challenge is inferring preferences from limited history rather than the method itself. In summary, APM provides a calibrated evaluation methodology that avoids judge biases for the study of implicit style personalization under realistic constraints.

\begin{ack}
This work was partially funded by the ERC (853489 - DEXIM) and the Alfried Krupp von Bohlen und Halbach Foundation, for which we thank them for their generous support. The authors gratefully acknowledge the scientific support and resources of the AI service infrastructure \textit{LRZ AI Systems} provided by the Leibniz Supercomputing Centre (LRZ) of the Bavarian Academy of Sciences and Humanities (BAdW), funded by Bayerisches Staatsministerium für Wissenschaft und Kunst (StMWK).
The authors gratefully acknowledge the Gauss Centre for Supercomputing e.V. (www.gauss-centre.eu) for funding this project by providing computing time on the GCS Supercomputer JUPITER at Jülich Supercomputing Centre (JSC).
We also acknowledge the use of the HPC cluster at Helmholtz Munich for the computational resources used in this study.


\end{ack}

\bibliographystyle{plainnat}
\bibliography{references}

\newpage
\appendix
\part*{Supplementary Material}

\section{Limitations and Broader Impact}
\label{sec:supp:limitations}

\mypara{Limitations}
Our evaluation relies on synthetic data, e.g.\ conversation histories are LLM-generated (by rewriting UltraChat prompts). This is necessary as no large-scale dataset of style-annotated multi-user conversations exists, and collecting real user judgments for the systematic experiments is prohibitively expensive at our scale. Also, we assume for APM that the set of response principles $\mathcal{P}$ is specified in advance, which constrains expressivity to predefined stylistic axes. This is a deliberate design choice that enables decomposing holistic ratings into the narrow per-principle judgments on which the unbiased baseline depends. Future work can attempt to relax this constraint and adapt to arbitrary or flexible stylistic axes.
In our experiments, we evaluate two open-weight base LLMs (\texttt{Llama-3.1-8B} and \texttt{Qwen-3.5-27B}), so our observations agree for models of different families and sizes. Nonetheless, models at entirely different scales or of different architectures may exhibit different behavior. Finally, all evaluated personalization methods cause small drops in general capabilities (MMLU-Pro, TruthfulQA) and demographic fairness (BBQ) relative to the non-personalized baseline (\cref{tab:main}). Reducing these side effects is left to future work.

\mypara{Broader Impact}
APM makes the evaluation of style personalization more reliable by separating genuine preference inference from the exploitation of LLM-judge biases and stereotypical attribute-style associations. This reduces the risk of deploying systems that appear personalized but actually reproduce a generic default style. Style personalization improves accessibility and user experience by adapting tone, length, or formality to individual users. Unlike content personalization, it does not directly affect the information users receive, which lowers concerns regarding filter bubbles or selective exposure.
At the same time, all evaluated methods cause small but consistent drops in demographic fairness on BBQ (e.g., $-0.141$ for routing on \texttt{Qwen-3.5-27B} in \cref{tab:main}). This indicates that style conditioning can interact with bias-related behavior and, like all AI systems, should be monitored before and during deployment. Furthermore, LLMs can manipulate users in some cases, and this risk potentially increases with strong personalization systems. We recommend using adequate guardrails in practice to detect and prevent malicious behavior or use.

\section{Empirical Justification of the Per-Principle Judge}

\label{app:category_based_scoring}
In \cref{sec:intro}, we argue that holistic LLM-judge personalization scoring can conflate adaptation to user preferences with general response quality and in particular rewards responses that just imitate the surface form of personalization, without any actual adjustment to the current user. Here, we provide evidence for this claim through an experiment that demonstrates both failure modes and contrasts them with the per-principle scoring, like we adopt in APM.

\paragraph{Setup.}
We sample 100 prompts from UltraChat and for each prompt we generate one response with \texttt{Qwen3.5-27B} using three different system prompts, and additionally we generate one response per prompt using \texttt{Qwen3.5-0.8B} without using any system prompt. The system-prompt conditions are: (i) no system prompt, (ii) a \textit{fake personalization} prompt instructing the model to write as if it knows the user well without providing any user information, and (iii) a \textit{warmth} prompt instructing the model to respond in a warm style. Each response is then rated by \texttt{gpt-oss-120b} on a 1--10 scale along four holistic criteria: overall quality, ungrounded personalization, personalization grounded on a randomly sampled attribute profile, and personalization grounded on a Nemotron-Personas-USA \citep{nvidia2025nemotron} persona. Additionally, we rate per-principle compliance for each of the 10 response principles in \cref{tab:attributes} and aggregate these into a \emph{sign-balanced principle score}, a single-response analog of the APM reward (\cref{eq:reward}). For each (prompt, principle) pair we take the judge's score and, with probability $0.5$, replace it with $11 - \text{score}$ before averaging across the 10 principles. This sign randomization mirrors the zero-mean construction of $\mathbf{C}$ in \cref{sec:formalization}: each principle is equally likely to be rewarded for adherence or opposition, so by the same argument as \cref{lemma:baseline}, any policy that does not condition on user information has expected score equal to the scale midpoint of $5.5$, regardless of judge bias.

\paragraph{Results.}
\cref{tab:judge-bias} directly reveals the failure modes of \texttt{gpt-oss-120b} when applied as judge for personalized responses. First, judge scores exhibit a directional quality bias. Across all three personalization criteria, \texttt{Qwen3.5-27B} receives between $16$ and $30$ percent higher scores than \texttt{Qwen3.5-0.8B} despite the fact that neither model is conditioned on any user information, so the personalization signal is in fact confounded with overall response quality. Second, surface cues that imitate personalization (condition (ii)) inflate scores in the absence of any actual user information. Instructing \texttt{Qwen3.5-27B} to act as if it knows the user well raises the ungrounded personalization score by $369\%$ over the no-system-prompt baseline. Grounding the judge on attribute profiles or Nemotron personas reduces but does not eliminate this.

Both phenomena in effect obscure whether observed score differences reflect true personalization or just correlate with stylistic features the judge rewards independently of personalization. The sign-balanced principle score is invariant to both confounds by construction. Because half of the (prompt, principle) scores are inverted before averaging, any policy that does not condition on user information collapses to the neutral midpoint of $5.5$. The rightmost column of \cref{tab:judge-bias} confirms this, as scores for all four configurations lie within $\pm 2$ percent of the midpoint of the scale (5.5).

\begin{table}[h]
\centering
\small
\resizebox{\linewidth}{!}{%
\begin{tabular}{lccccc}
\toprule
Configuration & Quality & Pers.\ ungrounded & Pers.\ attr-grounded & Pers.\ persona-grounded & Principle score \\
\midrule
Qwen3.5-27B, no system prompt (baseline) & 8.06 & 1.90 & 4.87 & 0.81 & 5.52 \\
Qwen3.5-0.8B, no system prompt           & 4.98 (-38.3\%) & 1.55 (-18.0\%) & 3.41 (-30.0\%) & 0.68 (-16.0\%) & 5.40 (-2.2\%) \\
Qwen3.5-27B, fake personalization prompt & 6.46 (-19.9\%) & 8.89 (+369.0\%) & 5.48 (+12.5\%) & 1.78 (+119.9\%) & 5.42 (-1.8\%) \\
Qwen3.5-27B, warmth prompt               & 7.62 (-5.5\%) & 3.92 (+106.8\%) & 5.56 (+14.3\%) & 1.10 (+36.3\%) & 5.48 (-0.7\%) \\
\bottomrule
\end{tabular}
}%
\caption{Judge scores (1--10 scale) under four configurations, with percentage change relative to the \texttt{Qwen-3.5-27B} no-system-prompt baseline. Personalization columns correspond to judges given no grounding, a randomly sampled attribute profile, or a Nemotron persona, respectively. The last column reports the sign-balanced principle score.}
\label{tab:judge-bias}
\end{table}

\section{Judge Selection}
\label{sec:supp:judgeselection}

For our work, it is very important to select a suitable LLM judge model. However, existing judge benchmarks \citep{tan2024judgebench, paech2024judgemarkv2} primarily focus on general helpfulness and safety rather than the stylistic discrimination our framework requires. Therefore, we conduct a custom comparison across a range of medium to large models, focusing on two properties that are easy to measure and directly relevant to the APM benchmark.

\begin{figure}[h]
    \centering
    \includegraphics[width=\textwidth]{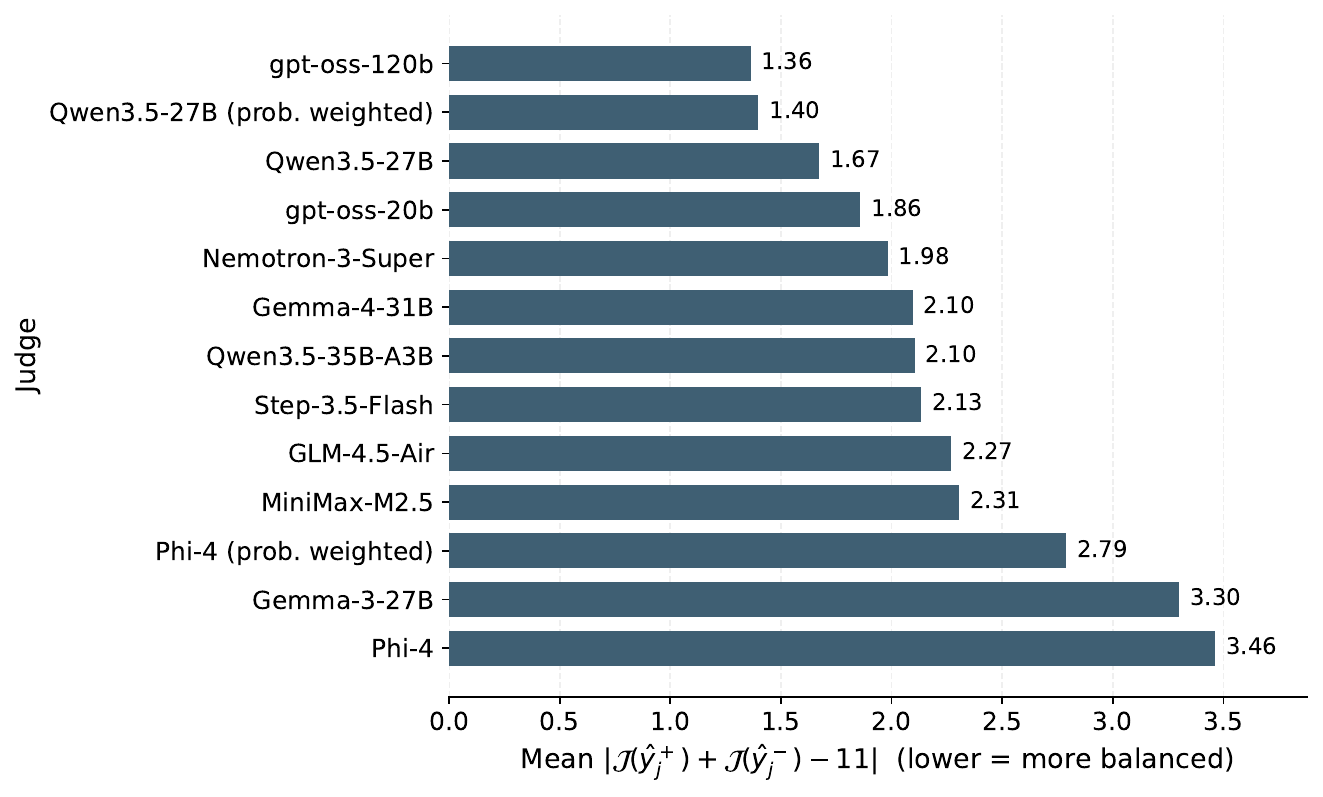}
    \caption{Mean L1 distance of $s^+ + s^-$ from $11$ per judge (lower is better).}
    \label{fig:judge_balance}
\end{figure}

\begin{figure}[h]
    \centering
    \includegraphics[width=\textwidth]{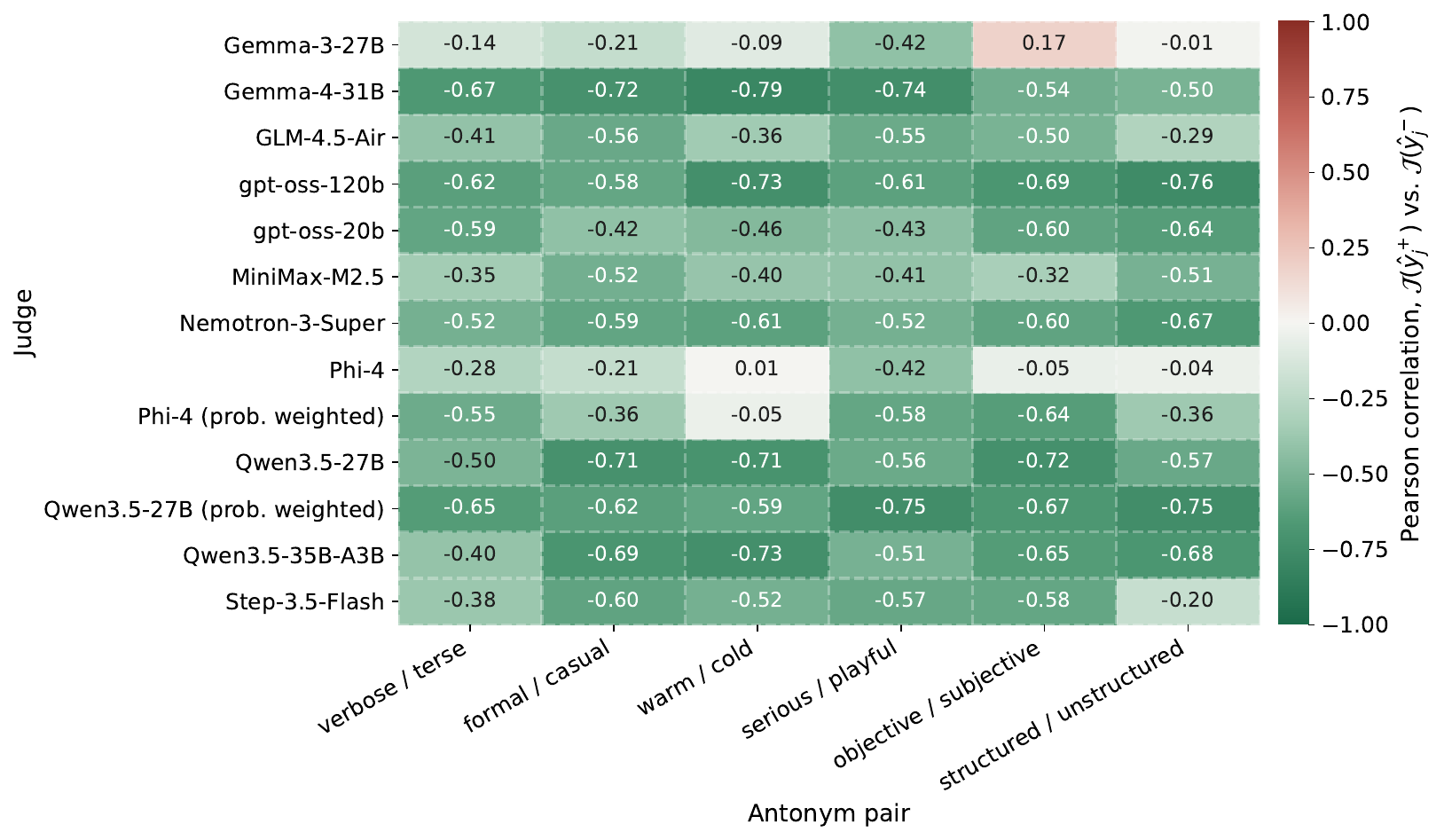}
    \caption{Pearson correlation between antonym scores per principle pair and judge. More negative values (green) indicate better discrimination.}
    \label{fig:judge_heatmap}
\end{figure}

\mypara{Evaluated properties}
For each candidate judge, we score a shared set of unaltered UltraChat responses against all response principles in \cref{tab:attributes}. We write $s^+$ to refer to the score when instructing the judge to evaluate whether the response follows the principle, and $s^-$ to refer to the score when we instruct the judge to evaluate whether the response avoids the principle. We measure:
\begin{enumerate}
    \item \textbf{Balance:} Whether the sum $s^+ + s^-$ is approximately constant ($\approx 11$ on our scale). A judge that assigns high scores to both directions of a principle injects systematic noise into training labels. We report the mean L1 distance $\mathbb{E}[|s^+ + s^- - 11|]$ (\cref{fig:judge_balance}).
    \item \textbf{Anticorrelation:} Whether $s^+$ and $s^-$ are negatively correlated across responses. A correlation near $-1$ indicates reliable discrimination between poles, and near $0$ suggests the judge conflates them. We report per-pair Pearson correlations (\cref{fig:judge_heatmap}).
\end{enumerate}

\mypara{Selection}
Among the evaluated models, \texttt{gpt-oss-120b} shows the best trade-off, as it achieves consistently low L1 divergence (Balance) and strong negative correlations while being significantly faster to run than \texttt{Qwen-27B} or \texttt{Qwen-35B-A3B} (the only models with similarly suitable properties) on our inference setup. Therefore, we adopt it as the default judge for all main experiments.

\section{Full Experimental Details}
\label{sec:supp:experimentaldetails}

\mypara{Experimental Setup}
We use \texttt{NVIDIA-Nemotron-3-Super} \citep{nvidia_nemotron_3_2025} as the base generation model for producing candidate responses across all personalization systems. Conversation histories are generated by prompting \texttt{gpt-oss-120b} \citep{openai2025gptoss120bgptoss20bmodel} to rewrite neutral prompts from UltraChat \citep{ding2023ultrachat} according to each user's assigned attributes. Unless otherwise stated, experiments use the following default configuration: $N=10$ personality dimensions and $M=10$ response principles, 10 signed permutation mapping matrices $\mathbf{C}$ (corresponding to 10 separate evaluations), one active trait per user ($k=1$), 4000 training users, 1000 test users, and 2 conversation history turns per user. As LLM Judge model, we use \texttt{gpt-oss-120b} \citep{openai2025gptoss120bgptoss20bmodel}, and prompt it to rate principle adherence on a scale from 1 to 10. As base LLMs in the personalization methods, we evaluate \texttt{Llama-3.1-8B} \citep{grattafiori2024llama} and \texttt{Qwen-3.5-27B} \citep{qwen3.5} to test models of different families and sizes.

\mypara{Method Hyperparameters}
All generation runs use vLLM in bf16 with vLLM's default sampling
($T{=}1.0$, top-$p{=}1.0$). The judge is run with $T{=}1.0$,
top-$p{=}0.95$, and thinking enabled. Method-specific hyperparameters
are listed in \cref{tab:hyperparams}. The oracle and unpersonalized
baselines have no trainable parameters and are omitted. LoRA is applied
to the attention projections \texttt{\{q,k,v,o\}\_proj} of the base
LLM.

\begin{table}[h]
\centering
\small
\caption{Hyperparameters for each personalization method.}
\label{tab:hyperparams}

\resizebox{\linewidth}{!}{%
\begin{tabular}{@{}c@{\hspace{1.5em}}c@{\hspace{1.5em}}c@{}}
\textbf{Routing} & \textbf{User-embedding (DPO)} & \textbf{RAG} \\[2pt]
\begin{tabular}[t]{ll}
\toprule
Backbone        & ModernBERT-base \\
Pooling         & mean \\
Precision       & fp32 \\
Loss            & cross-entropy \\
Optimizer       & AdamW \\
Learning rate   & $2{\times}10^{-5}$ \\
Eff.\ batch     & $4 \times 4 = 16$ \\
Epochs          & 3 \\
\bottomrule
\end{tabular}
&
\begin{tabular}[t]{ll}
\toprule
Backbone        & base LLM + LoRA \\
Encoder         & BGE-large (frozen) \\
LoRA $r/\alpha$/drop & $16/32/0.05$ \\
Soft tokens     & 1 (MLP projector) \\
Precision       & bf16 \\
Loss            & DPO ($\beta{=}0.1$) \\
Optimizer       & AdamW \\
Learning rate   & $5{\times}10^{-5}$ \\
Eff.\ batch     & $1 \times 8 = 8$ \\
Epochs          & 1 \\
\bottomrule
\end{tabular}
&
\begin{tabular}[t]{ll}
\toprule
Encoder       & BGE-large (frozen) \\
Retrieved $k$ & 3 \\
\bottomrule
\end{tabular}
\\
\end{tabular}}
\end{table}

\mypara{Computational Resources}
Experiments were conducted on a mix of NVIDIA H200, H100, and A100 GPUs. We optimize our setup by pre-generating per-user prompts and personalized responses according to all principles and embedding them. Then, training a single router takes approx.\ 1h on a single H100 GPU and learning soft prompts with associated LoRAs takes approx.\ 2 GPU-h, which is in line with standard LLM fine-tuning compute costs.

\section{User Attribute and Response Principle Selection}
\label{app:attribute_selection}

To use our evaluation, concrete user attributes and response principles need to be specified. In practice, they can be chosen depending on the use case or domain. In this work, we select suitable values using a data-driven method to demonstrate our evaluation framework. The pipeline below is designed to remove three distinct kinds of noise from a broad candidate pool: attributes that carry no discriminative signal (entropy filter), attributes that are redundant with each other (PCA), and components whose loadings are too diffuse to interpret as a single trait (Varimax rotation). We also explain how we create user-specific prompts based on the sampled attribute vector $\mathbf{a}_u$.

To select suitable attributes and principles, we begin with a broad candidate pool of values, which was generated by LLMs and manually curated to remove near-duplicates and attributes not directly related to style personalization. To select a suitable subset, we first score approximately 100 attributes on a set of 10,000 prompts and 100 principles on a set of 10,000 responses taken from UltraChat \citep{ding2023ultrachat}.

We compute the Shannon entropy of each attribute's score distribution (LLM judge scores) and discard attributes with entropy below 1.5. Low entropy means the judge assigns nearly the same score to almost every text, so the attribute carries no discriminative signal. After that, we compute the correlation matrix of the remaining attributes and use Parallel Analysis \citep{horn1965rationale} to estimate an upper bound $k_{\max}$ on the number of distinct latent dimensions. Parallel Analysis compares the eigenvalues of the observed correlation matrix to those obtained from random data of the same shape, retaining only components whose eigenvalues exceed what would be expected by chance. This acts as a noise floor and prevents us from modeling spurious structure.

We perform Principal Component Analysis on this correlation matrix, followed by a Varimax rotation \citep{kaiser1958varimax}, which concentrates each component's loading onto a small number of attributes. This sparsity allows us to approximate each latent dimension by its single highest-loading attribute, which the judge can then score directly. We retain a selected $k \leq k_{\max}$ components (we use $k=10$), by selecting the highest-loading attributes or principles from each component. The resulting values are in \cref{tab:attributes}.

\begin{table}[t]
\centering
\begin{tabular}{@{}llcll@{}}
\toprule
\multicolumn{2}{c}{\textbf{User Attributes}} & & \multicolumn{2}{c}{\textbf{Response Principles}} \\
\cmidrule(r){1-2} \cmidrule(l){4-5}
Verbose      & Transactional      & & Persuasive  & Acknowledges uncertainty       \\
Enthusiastic & Anxious            & & Elaborate           & Emotionally neutral                    \\
Prescriptive & Highly constrained & & Casual              & Assumes no prior knowledge     \\
Unpolished   & Tech-savvy         & & Prescriptive        & Collaborative                  \\
Open-ended   & Hypothetical       & & Gentle              & Presents multiple perspectives \\
\bottomrule
\end{tabular}
\caption{Selected user attributes and response principles. The 10 user attributes and response principles used in our experiments are derived from a large candidate pool via filtering and PCA. Final attributes and principles are summarized from the most relevant PCA loadings.}
\label{tab:attributes}
\end{table}

\Cref{alg:attribute_selection} gives the full procedure. We score 
$|\mathcal{A}| = 97$ candidate prompt attributes and $112$ candidate response principles on $n = 10{,}000$ UltraChat conversations, using $T = 100$ Gaussian surrogate datasets for Parallel Analysis (95th percentile threshold). The entropy filter retains $67/97$ prompt attributes and $88/112$ response principles, indicating that roughly a quarter of LLM-proposed candidates carry no discriminative signal under judge scoring. Parallel Analysis yields $k_{\max} = 10$ for prompt attributes and $k_{\max} = 12$ for response principles, with the top-$k_{\max}$ components explaining $63.2\%$ and $61.9\%$ of total variance. We select $k = 10$ in both cases. Resulting components and their highest-loading attributes are in \Cref{tab:user_attributes,tab:response_attributes}.

\begin{algorithm}[hbt!]
\caption{Algorithm for Attribute Selection}
\label{alg:attribute_selection}
\begin{algorithmic}[1]
\Require Texts $D$, Candidate Attributes $A$, LLM Judge $J$, Entropy Threshold $\tau$, Target Dimensions $k$
\Ensure A disentangled subset of attributes $A_{\text{selected}}$ where $|A_{\text{selected}}| = k$

\State \textbf{Initialize} score matrix $\mathbf{S} \in \mathbb{R}^{|D| \times |A|}$
\For{each text $d \in D$ and attribute $a \in A$}
    \State $\mathbf{S}[d, a] \leftarrow J(d, a)$ \Comment{Score text $d$ on attribute $a$}
\EndFor

\State $A_{\text{filtered}} \leftarrow \emptyset$
\For{each attribute $a \in A$}
    \State $p \leftarrow \text{Distribution}(\mathbf{S}[:, a])$
    \If{$-\sum p \log_2 p \geq \tau$}
        \State $A_{\text{filtered}} \leftarrow A_{\text{filtered}} \cup \{a\}$
    \EndIf
\EndFor

\State $\mathbf{Z} \leftarrow \text{StandardizeColumns}(\mathbf{S}[:, A_{\text{filtered}}])$
\State $\mathbf{C} \leftarrow \frac{1}{|D|-1} \mathbf{Z}^\top \mathbf{Z}$ \Comment{Compute Correlation Matrix}
\State $\mathbf{\Lambda}, \mathbf{V} \leftarrow \text{Eigendecomposition}(\mathbf{C})$ \Comment{Eigenvalues and Eigenvectors}

\State $\mathbf{\Lambda}_{\text{rand}} \leftarrow \text{SimulateRandomEigenvalues}(|D|, |A_{\text{filtered}}|)$
\State $k_{\max} \leftarrow \sum \mathbb{I}(\mathbf{\Lambda} > \text{Percentile}(\mathbf{\Lambda}_{\text{rand}}, 95))$ \Comment{Parallel Analysis}
\State \textbf{Assert} $k \leq k_{\max}$

\State $\mathbf{L} \leftarrow \mathbf{V}[:, 1\dots k] \times \text{diag}(\sqrt{\mathbf{\Lambda}_{1\dots k}})$ \Comment{Extract initial loadings}
\State $\mathbf{L}_{\text{rot}} \leftarrow \text{VarimaxRotation}(\mathbf{L})$ \Comment{Force sparse, interpretable loadings}

\State $A_{\text{selected}} \leftarrow \emptyset$
\For{$j = 1$ \textbf{to} $k$}
    \State $a^* \leftarrow \arg\max_{a \in A_{\text{filtered}}} |\mathbf{L}_{\text{rot}}[a, j]|$ \Comment{Select highest loading attribute}
    \State $A_{\text{selected}} \leftarrow A_{\text{selected}} \cup \{a^*\}$
\EndFor

\State \Return $A_{\text{selected}}$
\end{algorithmic}
\end{algorithm}

\begin{table}[h]
\centering
\caption{Top 5 attribute loadings for the 10 Rotated Components (User Prompt Attributes). The representative attribute for each component is the one with the highest absolute loading. SSL = Sum of Squared Loadings.}
\label{tab:user_attributes}
\small
\begin{tabular}{@{}llcccc@{}}
\toprule
\textbf{Component} & \textbf{Attribute} & \textbf{Loading} & \textbf{Mean} & \textbf{Std} & \textbf{Entropy} \\
\midrule
\textbf{RC1 (Verbose)} & verbose & +0.877 & 3.71 & 2.45 & 2.649 \\
\textit{SSL: 10.486} & simple language & -0.850 & 7.13 & 2.24 & 2.683 \\
& wall-of-text & +0.840 & 2.33 & 2.42 & 1.880 \\
& sophisticated vocabulary & +0.794 & 3.10 & 1.52 & 2.234 \\
& terse & -0.794 & 5.22 & 2.77 & 3.080 \\
\midrule
\textbf{RC2 (Enthusiastic)} & enthusiastic & -0.863 & 4.36 & 2.39 & 2.804 \\
\textit{SSL: 8.898} & dry & +0.797 & 5.66 & 2.88 & 3.065 \\
& neutral & +0.787 & 7.62 & 2.97 & 2.622 \\
& colloquial & -0.749 & 4.46 & 2.42 & 2.923 \\
& formal & +0.746 & 5.70 & 2.43 & 2.928 \\
\midrule
\textbf{RC3 (Prescriptive)} & prescriptive & -0.725 & 4.56 & 3.34 & 2.814 \\
\textit{SSL: 4.979} & commanding & -0.642 & 2.71 & 1.96 & 1.988 \\
& polite & +0.586 & 8.12 & 1.80 & 2.415 \\
& fact-seeking & +0.574 & 7.17 & 3.58 & 2.447 \\
& rule-heavy & -0.543 & 2.23 & 2.11 & 1.963 \\
\midrule
\textbf{RC4 (Unpolished)} & unpolished & -0.728 & 2.34 & 1.77 & 2.011 \\
\textit{SSL: 4.111} & articulate & +0.712 & 8.24 & 1.10 & 1.717 \\
& meticulously edited & +0.621 & 5.62 & 2.19 & 2.847 \\
& rambling & -0.579 & 2.00 & 1.88 & 1.795 \\
& earnest & +0.497 & 7.63 & 1.64 & 2.130 \\
\midrule
\textbf{RC5 (Open-ended)} & open-ended & +0.705 & 7.11 & 2.58 & 2.784 \\
\textit{SSL: 3.527} & brainstorming & +0.664 & 3.37 & 2.44 & 2.593 \\
& flexible & +0.643 & 4.77 & 2.52 & 2.754 \\
& open-minded & +0.632 & 6.24 & 2.26 & 2.719 \\
& exploratory & +0.631 & 6.53 & 2.48 & 2.837 \\
\midrule
\textbf{RC6 (Transactional)} & transactional & +0.638 & 6.18 & 3.26 & 2.852 \\
\textit{SSL: 2.462} & collaborative & +0.598 & 4.69 & 2.88 & 2.903 \\
& trusting & +0.496 & 4.40 & 2.53 & 2.858 \\
& patient & +0.450 & 5.14 & 2.40 & 2.895 \\
& sycophantic & +0.380 & 1.88 & 1.46 & 1.826 \\
\midrule
\textbf{RC7 (Anxious)} & anxious & +0.755 & 1.64 & 1.05 & 1.540 \\
\textit{SSL: 2.094} & troubleshooting & +0.602 & 2.03 & 1.77 & 1.894 \\
& guarded & +0.446 & 1.95 & 1.26 & 1.739 \\
& validation-seeking & +0.407 & 2.22 & 1.65 & 2.009 \\
& confessional & +0.399 & 1.83 & 1.58 & 1.715 \\
\midrule
\textbf{RC8 (Highly constrained)} & highly constrained & +0.559 & 5.44 & 2.41 & 3.077 \\
\textit{SSL: 1.957} & vague & -0.552 & 2.10 & 1.47 & 1.653 \\
& precise & +0.513 & 6.73 & 2.20 & 2.626 \\
& high-context & +0.347 & 4.40 & 2.69 & 2.730 \\
& fact-seeking & +0.299 & 7.17 & 3.58 & 2.447 \\
\midrule
\textbf{RC9 (Tech-savvy)} & tech-savvy & +0.763 & 2.82 & 2.29 & 2.456 \\
\textit{SSL: 1.951} & jargon-heavy & +0.480 & 2.27 & 1.75 & 2.225 \\
& domain-expert & +0.474 & 3.82 & 2.18 & 2.685 \\
& analytical & +0.319 & 3.54 & 2.18 & 2.532 \\
& guarded & +0.281 & 1.95 & 1.26 & 1.739 \\
\midrule
\textbf{RC10 (Hypothetical)} & hypothetical & +0.647 & 2.61 & 2.71 & 2.068 \\
\textit{SSL: 1.891} & scenario-building & +0.566 & 3.81 & 2.62 & 2.713 \\
& creative & +0.388 & 2.92 & 1.62 & 2.098 \\
& fact-seeking & -0.335 & 7.17 & 3.58 & 2.447 \\
& calm & -0.324 & 7.30 & 2.08 & 2.322 \\
\bottomrule
\end{tabular}
\end{table}

\begin{table}[h]
\centering
\caption{Top 5 attribute loadings for the 10 Rotated Components (LLM Response Attributes). The representative attribute for each component is the one with the highest absolute loading. SSL = Sum of Squared Loadings.}
\label{tab:response_attributes}
\small
\begin{tabular}{@{}llcccc@{}}
\toprule
\textbf{Component} & \textbf{Attribute} & \textbf{Loading} & \textbf{Mean} & \textbf{Std} & \textbf{Entropy} \\
\midrule
\textbf{RC1 (Emotionally neutral)} & emotionally neutral & +0.864 & 7.10 & 2.67 & 2.889 \\
\textit{SSL: 12.326} & appeals to emotion & -0.850 & 3.40 & 2.35 & 2.856 \\
& dry & +0.812 & 6.66 & 2.44 & 2.898 \\
& stoic & +0.809 & 5.55 & 2.83 & 3.016 \\
& clinical & +0.803 & 5.56 & 2.78 & 3.121 \\
\midrule
\textbf{RC2 (Elaborate)} & elaborate & +0.844 & 6.32 & 1.83 & 2.620 \\
\textit{SSL: 7.865} & verbose & +0.792 & 6.48 & 2.10 & 2.748 \\
& depth-oriented & +0.782 & 4.75 & 1.64 & 2.692 \\
& granular & +0.598 & 5.91 & 1.84 & 2.808 \\
& terse & -0.594 & 3.63 & 2.18 & 2.771 \\
\midrule
\textbf{RC3 (Casual)} & casual & +0.822 & 4.43 & 2.03 & 2.848 \\
\textit{SSL: 7.304} & colloquial & +0.780 & 3.34 & 1.68 & 2.477 \\
& peer-to-peer tone & +0.780 & 4.59 & 2.33 & 2.907 \\
& conversational & +0.740 & 5.17 & 2.06 & 2.862 \\
& playful & +0.641 & 2.11 & 1.16 & 1.863 \\
\midrule
\textbf{RC4 (Prescriptive)} & prescriptive & -0.840 & 4.80 & 3.19 & 3.008 \\
\textit{SSL: 6.074} & action-oriented & -0.813 & 5.51 & 2.65 & 3.135 \\
& gives specific recommendations & -0.774 & 4.78 & 3.51 & 2.914 \\
& step-by-step & -0.672 & 3.69 & 2.54 & 2.879 \\
& heavy formatting & -0.579 & 2.07 & 1.39 & 1.989 \\
\midrule
\textbf{RC5 (Gentle)} & gentle & +0.688 & 7.26 & 1.60 & 2.448 \\
\textit{SSL: 4.382} & calming & +0.654 & 5.18 & 1.81 & 2.853 \\
& diplomatic & +0.604 & 8.06 & 1.46 & 2.094 \\
& reassuring & +0.475 & 5.23 & 2.04 & 2.913 \\
& patient & +0.430 & 6.94 & 1.81 & 2.600 \\
\midrule
\textbf{RC6 (Acknowledges uncertainty)} & acknowledges uncertainty & -0.800 & 2.22 & 2.37 & 1.835 \\
\textit{SSL: 4.034} & hedged & -0.739 & 2.50 & 2.06 & 2.348 \\
& confident & +0.710 & 7.99 & 1.76 & 2.307 \\
& cautious & -0.644 & 4.21 & 2.39 & 2.814 \\
& challenges user assumptions & -0.526 & 1.83 & 1.40 & 1.683 \\
\midrule
\textbf{RC7 (Persuasive)} & persuasive & +0.605 & 4.44 & 1.93 & 2.809 \\
\textit{SSL: 2.803} & opinionated & +0.570 & 2.27 & 1.68 & 2.194 \\
& non-judgmental & -0.423 & 9.15 & 1.46 & 1.785 \\
& optimistic & +0.344 & 5.80 & 2.42 & 3.048 \\
& appeals to authority & +0.316 & 2.23 & 2.00 & 2.073 \\
\midrule
\textbf{RC8 (Assumes no prior knowledge)} & assumes no prior knowledge & -0.610 & 7.20 & 1.93 & 2.728 \\
\textit{SSL: 2.719} & jargon-heavy & +0.506 & 3.09 & 1.87 & 2.656 \\
& exhaustive & -0.440 & 5.49 & 1.98 & 2.967 \\
& concrete & -0.423 & 7.20 & 1.97 & 2.709 \\
& simple language & -0.406 & 7.19 & 1.28 & 2.287 \\
\midrule
\textbf{RC9 (Collaborative)} & collaborative & +0.602 & 4.52 & 2.45 & 2.982 \\
\textit{SSL: 2.384} & defers to the user & +0.494 & 6.08 & 3.42 & 2.872 \\
& deferential & +0.486 & 5.53 & 2.50 & 3.065 \\
& validating & +0.465 & 5.73 & 2.46 & 3.096 \\
& service-oriented & +0.362 & 7.75 & 1.84 & 2.419 \\
\midrule
\textbf{RC10 (Presents multiple perspectives)} & presents multiple perspectives & +0.525 & 3.31 & 2.18 & 2.689 \\
\textit{SSL: 2.164} & commits to one answer & -0.494 & 7.70 & 3.00 & 2.261 \\
& breadth-oriented & +0.475 & 6.37 & 2.09 & 2.731 \\
& speaks in absolutes & -0.450 & 2.49 & 2.06 & 2.318 \\
& example-driven & +0.431 & 6.31 & 2.66 & 3.154 \\
\bottomrule
\end{tabular}
\end{table}

\section{Lexical Analysis of Rewritten Prompts}
\label{sec:supp:promptrewriting}

To simulate user interactions, we construct the conversation history $\mathcal{H}_u$ using an LLM based on sampled user attribute vectors $\mathbf{a}_u$. We sample prompts from the UltraChat dataset \citep{ding2023ultrachat}, then instruct an LLM (\texttt{gpt-oss-120b} \citep{openai2025gptoss120bgptoss20bmodel} in our case) to rewrite these prompts from the perspective of the user's sampled personality attributes $\mathbf{a}_u$. This produces a per-user conversation history $\mathcal{H}_u$ that implicitly reflects the user's personality in the way the prompts are formulated.

To verify that the rewriting procedure produces stylistically distinct prompts along each attribute axis, we perform a bigram analysis on the training set. For each attribute $a_i$ and each side (follow / avoid), we rank bigrams by their KL contribution $P(g \mid \text{side}) \log \tfrac{P(g \mid \text{side})}{P(g \mid \text{rest})}$, where the comparison set excludes both sides of $a_i$ to isolate the attribute's effect. This metric favors bigrams that are both frequent on the target side and disproportionately rare elsewhere. Results are shown in \cref{tab:2_grams}. The top bigrams on the follow side are semantically aligned with the intended attribute, and the avoid side surfaces its own coherent register rather than reading as stylistically neutral.

\begin{table}[t]
\centering
\resizebox{\linewidth}{!}{%
\begin{tabular}{@{}l p{7cm} p{7cm}@{}}
\toprule
\textbf{Attribute} & \textbf{Follow side} & \textbf{Avoid side} \\
\midrule
anxious & nervous about (kl=2.40, 6341.9$\times$), bit nervous (kl=1.94, 16469.7$\times$), anxious about (kl=1.71, 8409.3$\times$), you please (kl=1.57, 85.7$\times$), but could (kl=1.16, 1198.8$\times$), could you (kl=1.13, 4.4$\times$), about getting (kl=1.04, 783.0$\times$), feeling bit (kl=1.04, 2338.7$\times$) & clear step (kl=0.06, 5.9$\times$), provide clear (kl=0.05, 6.2$\times$), explain how (kl=0.04, 1.8$\times$), concise summary (kl=0.04, 2.7$\times$), me clear (kl=0.03, 10.5$\times$), based on (kl=0.02, 1.4$\times$), outline the (kl=0.02, 2.1$\times$), summary of (kl=0.02, 1.6$\times$) \\
\addlinespace
enthusiastic & hey there (kl=0.86, 140.4$\times$), excited to (kl=0.76, 284.6$\times$), thrilled to (kl=0.56, 427.8$\times$), dive into (kl=0.41, 33.0$\times$), there could (kl=0.39, 114.2$\times$), an exciting (kl=0.31, 171.0$\times$), super excited (kl=0.28, 2292.8$\times$), eager to (kl=0.27, 42.6$\times$) & provide straightforward (kl=0.09, 53.4$\times$), such as (kl=0.05, 1.9$\times$), what is (kl=0.03, 3.6$\times$), suitable for (kl=0.03, 4.1$\times$), of the (kl=0.03, 1.2$\times$), provide factual (kl=0.03, 102.0$\times$), write straightforward (kl=0.03, 24.0$\times$), is the (kl=0.02, 2.3$\times$) \\
\addlinespace
highly constrained & provide concise (kl=0.10, 3.5$\times$), write concise (kl=0.07, 6.9$\times$), concise step (kl=0.07, 6.6$\times$), summarize the (kl=0.05, 2.8$\times$), list the (kl=0.04, 2.4$\times$), write brief (kl=0.03, 7.2$\times$), create concise (kl=0.03, 5.8$\times$), no more (kl=0.02, 7.9$\times$) & hey could (kl=0.20, 13.0$\times$), could you (kl=0.13, 1.5$\times$), put together (kl=0.10, 4.5$\times$), can you (kl=0.09, 2.5$\times$), you put (kl=0.08, 5.3$\times$), you give (kl=0.06, 2.4$\times$), give me (kl=0.06, 2.0$\times$), rundown of (kl=0.06, 4.2$\times$) \\
\addlinespace
hypothetical & imagine you (kl=1.74, 390.8$\times$), suppose you (kl=0.60, 12317.2$\times$), were to (kl=0.55, 509.5$\times$), you re (kl=0.52, 16.9$\times$), you need (kl=0.32, 26.1$\times$), if we (kl=0.32, 77.2$\times$), re tasked (kl=0.31, 18179.0$\times$), you were (kl=0.30, 50.6$\times$) & based on (kl=0.03, 1.5$\times$), you summarize (kl=0.02, 6.3$\times$), the passage (kl=0.02, 1.3$\times$), explain how (kl=0.02, 1.4$\times$), such as (kl=0.02, 1.3$\times$), on the (kl=0.02, 1.1$\times$), you outline (kl=0.01, 2.6$\times$), what ways (kl=0.01, 2.9$\times$) \\
\addlinespace
open-ended & could you (kl=0.36, 2.3$\times$), curious about (kl=0.13, 6.8$\times$), you walk (kl=0.12, 5.9$\times$), walk me (kl=0.12, 5.3$\times$), me through (kl=0.12, 5.0$\times$), you explore (kl=0.08, 15.8$\times$), interested in (kl=0.07, 4.8$\times$), feel free (kl=0.07, 5.1$\times$) & list three (kl=0.08, 114.8$\times$), list the (kl=0.07, 3.1$\times$), three specific (kl=0.04, 21.1$\times$), summarize the (kl=0.03, 2.2$\times$), identify three (kl=0.03, 151.1$\times$), specific ways (kl=0.03, 9.1$\times$), based on (kl=0.02, 1.4$\times$), explain how (kl=0.02, 1.4$\times$) \\
\addlinespace
prescriptive & provide detailed (kl=0.24, 11.1$\times$), follow these (kl=0.21, 108.4$\times$), these exact (kl=0.18, 826.5$\times$), by step (kl=0.14, 2.4$\times$), step by (kl=0.14, 2.4$\times$), the specific (kl=0.13, 5.9$\times$), ensure the (kl=0.12, 4.8$\times$), of the (kl=0.12, 1.6$\times$) & could you (kl=0.27, 2.0$\times$), curious about (kl=0.22, 10.6$\times$), looking for (kl=0.16, 5.6$\times$), interested in (kl=0.15, 8.7$\times$), would be (kl=0.15, 5.4$\times$), it would (kl=0.13, 11.3$\times$), be great (kl=0.10, 13.8$\times$), you share (kl=0.10, 4.7$\times$) \\
\addlinespace
tech-savvy & data driven (kl=0.42, 53.5$\times$), tech savvy (kl=0.29, 190.5$\times$), parse the (kl=0.19, 499.7$\times$), tech focused (kl=0.15, 519.7$\times$), concise tech (kl=0.12, 1898.4$\times$), high resolution (kl=0.10, 17.0$\times$), tech oriented (kl=0.09, 5889.7$\times$), tech driven (kl=0.07, 127.2$\times$) & can you (kl=0.36, 6.5$\times$), me simple (kl=0.17, 98.9$\times$), you give (kl=0.17, 4.5$\times$), easy to (kl=0.16, 4.9$\times$), give me (kl=0.14, 3.1$\times$), plain language (kl=0.13, 28.7$\times$), an easy (kl=0.11, 22.3$\times$), and how (kl=0.10, 2.2$\times$) \\
\addlinespace
transactional & provide concise (kl=0.27, 7.2$\times$), list the (kl=0.07, 3.3$\times$), concise step (kl=0.07, 6.8$\times$), provide the (kl=0.06, 7.0$\times$), please provide (kl=0.05, 4.5$\times$), concise summary (kl=0.04, 2.6$\times$), provide list (kl=0.03, 10.0$\times$), list of (kl=0.03, 1.9$\times$) & could you (kl=0.35, 2.2$\times$), hey could (kl=0.17, 11.3$\times$), you walk (kl=0.13, 6.0$\times$), walk me (kl=0.12, 5.4$\times$), me through (kl=0.12, 5.2$\times$), curious about (kl=0.09, 5.3$\times$), curious how (kl=0.09, 12.3$\times$), curious what (kl=0.07, 11.6$\times$) \\
\addlinespace
unpolished & throw in (kl=0.59, 101.7$\times$), hey can (kl=0.46, 60.8$\times$), can you (kl=0.41, 6.8$\times$), in some (kl=0.22, 35.7$\times$), toss in (kl=0.21, 41.5$\times$), give me (kl=0.18, 3.5$\times$), yo can (kl=0.18, 2595.5$\times$), talk about (kl=0.18, 18.6$\times$) & you provide (kl=0.09, 5.4$\times$), please provide (kl=0.06, 5.0$\times$), could you (kl=0.06, 1.2$\times$), such as (kl=0.06, 1.9$\times$), of the (kl=0.05, 1.3$\times$), ensure the (kl=0.05, 2.7$\times$), you kindly (kl=0.05, 5.3$\times$), kindly provide (kl=0.04, 11.6$\times$) \\
\addlinespace
verbose & an extensive (kl=0.80, 394.5$\times$), of the (kl=0.65, 3.6$\times$), compose an (kl=0.55, 47.7$\times$), would greatly (kl=0.52, 718.7$\times$), greatly appreciate (kl=0.50, 588.1$\times$), an exhaustive (kl=0.45, 238.1$\times$), the myriad (kl=0.42, 258.5$\times$), richly detailed (kl=0.40, 84.2$\times$) & write concise (kl=0.09, 8.3$\times$), summarize the (kl=0.05, 2.8$\times$), write brief (kl=0.05, 10.7$\times$), concise step (kl=0.03, 4.0$\times$), how do (kl=0.03, 2.5$\times$), brief step (kl=0.03, 30.1$\times$), give brief (kl=0.03, 6.4$\times$), write 500 (kl=0.03, 3.6$\times$) \\
\bottomrule
\end{tabular}
}
\caption{Top bigrams per attribute on the follow and avoid sides, ranked by KL contribution against the rest of the dataset (i.e., excluding both sides of the same attribute). Each entry shows the bigram, its KL contribution, and the frequency ratio to the comparison set.}
\label{tab:2_grams}
\end{table}

\section{Proof of Baseline Win-Rate}
\label{sec:supp:baselinelemma}

Here, we prove that the baseline win-rate is exactly $50\%$ under a mild symmetry condition on the sampling distribution of $\mathbf{C}$, which complements the zero-reward result in \cref{sec:formalization}. We restate \cref{lemma:baseline} for convenience.

\mypara{Theorem \ref{lemma:baseline} (restated)}
Let $\pi_0$ be a policy that does not condition on the user's history $\mathcal{H}_u$, and let $y_0 = \pi_0(x)$. Assume:
\begin{enumerate}[label=(\roman*), leftmargin=*]
\item $\mathbb{E}[\mathbf{C}_{ji}] = 0$ for all $(j, i)$;
\item $\mathbf{C}$ is independent of $(\mathbf{a}_u, x, y_0)$.
\end{enumerate}
Then $\mathbb{E}[r(u, x, \pi_0)] = 0$ for any judge $\mathcal{J}$.

The win-rate $\mathbb{E}[W]$ is computed by comparing a candidate response $y_A$ against a reference response $y_B$ for each user-prompt pair $(u, x)$. $y_A$ wins if its reward under the user preferences is strictly greater than the reward of $y_B$. To formalize this, we define:
\begin{align}
W &= \mathbb{I}[r(u, x, y_A) > r(u, x, y_B)] \\
\mathbf{d} &= \mathcal{J}(y_A) - \mathcal{J}(y_B) \in \mathbb{R}^M \\
S &= r(u, x, y_A) - r(u, x, y_B) = \mathbf{a}^T_u\mathbf{C}^T\mathbf{d} = (\mathbf{C}\mathbf{a}_u)^T\mathbf{d}
\end{align}

We note that $W = \mathbb{I}[S > 0]$. We now state the additional theorem we want to prove:

\begin{theorem}[Baseline win-rate]
\label{cor:winrate}
Let $y_A$ and $y_B$ both come from policies that do not condition on $\mathcal{H}_u$. Assume:
\begin{enumerate}[label=(\roman*), leftmargin=*]
\item $\mathbf{C}$ is independent of $(\mathbf{a}_u, x, y_A, y_B)$;
\item conditional on $\mathbf{a}_u$, $\mathbf{C}\mathbf{a}_u \stackrel{d}{=} -\mathbf{C}\mathbf{a}_u$.
\end{enumerate}
Then $S \stackrel{d}{=} -S$, and consequently $\mathbb{E}[W] \;=\; \tfrac{1}{2}\bigl(1 - P(S = 0)\bigr)$.
If ties are broken uniformly at random, or equivalently counted as half-wins via $\widetilde{W} = \mathbb{I}[S > 0] + \tfrac{1}{2}\mathbb{I}[S = 0]$, then $\mathbb{E}[\widetilde{W}]$ equals $\tfrac{1}{2}$ exactly.
\end{theorem}

\mypara{Proof}
The intuition is that, by assumption (ii), the preference vector $\mathbf{C}\mathbf{a}_u$ has the same distribution as its negation. Flipping its sign produces an equally likely scenario. Because $S = (\mathbf{C}\mathbf{a}_u)^\top \mathbf{d}$ is a linear function of this vector, $S$ inherits the same symmetry. We only need to consider that $\mathbf{d}$ depends on $\mathbf{a}_u$ in our setup, since prompts are rewritten in the user style. We handle this by conditioning on the value of $\mathbf{a}_u$ before we apply the symmetry.

We fix arbitrary realizations $\mathbf{a}_u = \mathbf{a}_0$ and $\mathbf{d} = \mathbf{d}_0$ and consider the distribution of $S$ given these values. With $\mathbf{a}_0$ and $\mathbf{d}_0$ treated as constants, $S$ reduces to a linear function of the random matrix $\mathbf{C}$:
\begin{equation}
S \,\big|\, (\mathbf{a}_u = \mathbf{a}_0,\, \mathbf{d} = \mathbf{d}_0) \;=\; \mathbf{d}_0^\top \mathbf{C} \mathbf{a}_0.
\end{equation}

We notice that this conditional distribution is symmetric about zero. First, by assumption (i), conditioning on $(\mathbf{a}_u, \mathbf{d})$ does not change the distribution of $\mathbf{C}$. Second, by assumption (ii) applied at the specific value $\mathbf{a}_u = \mathbf{a}_0$, the random vector $\mathbf{C}\mathbf{a}_0$ has the same distribution as $-\mathbf{C}\mathbf{a}_0$. Equality in distribution is preserved under multiplication by any fixed vector. Taking the inner product with $\mathbf{d}_0$ on both sides gives:
\begin{equation}
\mathbf{d}_0^\top (\mathbf{C}\mathbf{a}_0) \;\stackrel{d}{=}\; \mathbf{d}_0^\top (-\mathbf{C}\mathbf{a}_0) \;=\; -\mathbf{d}_0^\top (\mathbf{C}\mathbf{a}_0).
\end{equation}

Combining these points, we see that the conditional distribution of $S$ given any choice of $(\mathbf{a}_u, \mathbf{d})$ is symmetric about zero. Averaging over the joint distribution of $(\mathbf{a}_u, \mathbf{d})$ gives the unconditional statement $S \stackrel{d}{=} -S$.

It follows that $P(S > 0) = P(-S > 0) = P(S < 0)$. From $1 = P(S > 0) + P(S < 0) + P(S = 0)$, we obtain $P(S > 0) = \tfrac{1}{2}\bigl(1 - P(S = 0)\bigr)$, which is $\mathbb{E}[W]$. Counting ties as half-wins, we get $\mathbb{E}[\widetilde{W}] \;=\; P(S > 0) + \tfrac{1}{2} P(S = 0) \;=\; \tfrac{1}{2}$.

\end{document}